\documentclass[letterpaper]{article} 
\usepackage[submission]{aaai2026}  
\usepackage{times}  
\usepackage{helvet}  
\usepackage{courier}  
\usepackage[hyphens]{url}  
\usepackage{graphicx} 
\usepackage{natbib}  
\usepackage{caption} 
\usepackage{algorithm}
\usepackage{algorithmic}
\usepackage{multirow}
\usepackage[linesnumbered,ruled,vlined,algo2e]{algorithm2e}
\usepackage{subcaption}  
\usepackage{amsmath}
\usepackage{amsfonts}
\usepackage{xcolor}

\usepackage{newfloat}
\usepackage{listings}
\DeclareCaptionStyle{ruled}{labelfont=normalfont,labelsep=colon,strut=off} 
\floatstyle{ruled}
\newfloat{listing}{tb}{lst}{}
\floatname{listing}{Listing}
\usepackage[submission]{aaai2026}
\makeatletter
\gdef\showauthors@on{T}
\makeatother

\title{\textsc{NeuroLogic}: From Neural Representations to Interpretable Logic Rules}
\author {
    Chuqin Geng\textsuperscript{\rm 1},
    Anqi Xing\textsuperscript{\rm 2},
    Li Zhang\textsuperscript{\rm 2},
    Ziyu Zhao\textsuperscript{\rm 1},
    Yuhe Jiang\textsuperscript{\rm 2},
    Xujie Si\textsuperscript{\rm 2},
}
\affiliations {
    \textsuperscript{\rm 1}McGill University\\
    \textsuperscript{\rm 2}University of Toronto\\
}

\begin{document}

\maketitle

\begin{abstract}

Rule-based explanation methods offer rigorous and globally interpretable insights into neural network behavior. However, existing approaches are mostly limited to small fully connected networks and depend on costly layer-wise rule extraction and substitution processes. These limitations hinder their generalization to more complex architectures such as Transformers. Moreover, existing methods produce shallow, decision-tree-like rules that fail to capture rich, high-level abstractions in complex domains like computer vision and natural language processing. To address these challenges, we propose \textsc{NeuroLogic}, a novel framework that extracts interpretable logical rules directly from deep neural networks. Unlike previous methods, \textsc{NeuroLogic} can construct logic rules over hidden predicates derived from neural representations at any chosen layer, in contrast to costly layer-wise extraction and rewriting. This flexibility enables broader architectural compatibility and improved scalability. Furthermore, \textsc{NeuroLogic} supports richer logical constructs and can incorporate human prior knowledge to ground hidden predicates back to the input space, enhancing interpretability. We validate \textsc{NeuroLogic} on Transformer-based sentiment analysis, demonstrating its ability to extract meaningful, interpretable logic rules and provide deeper insights—tasks where existing methods struggle to scale.

\end{abstract}


%

\section{Introduction}
\label{sec:intro}
In recent years, deep neural networks have made remarkable progress across various domains, including computer vision~\cite{2012icwdc, 2016drlfi} and natural language processing~\cite{2014stslw}. As AI advances, the demand for interpretability has become increasingly urgent especially in high-stakes and regulated domains where understanding model decisions is critical~\cite{lipton2016tmomi, doshi2017tarso, guidotti2018asomf}.



\begin{figure*}[t] 
    \centering
    \includegraphics[width=\textwidth]{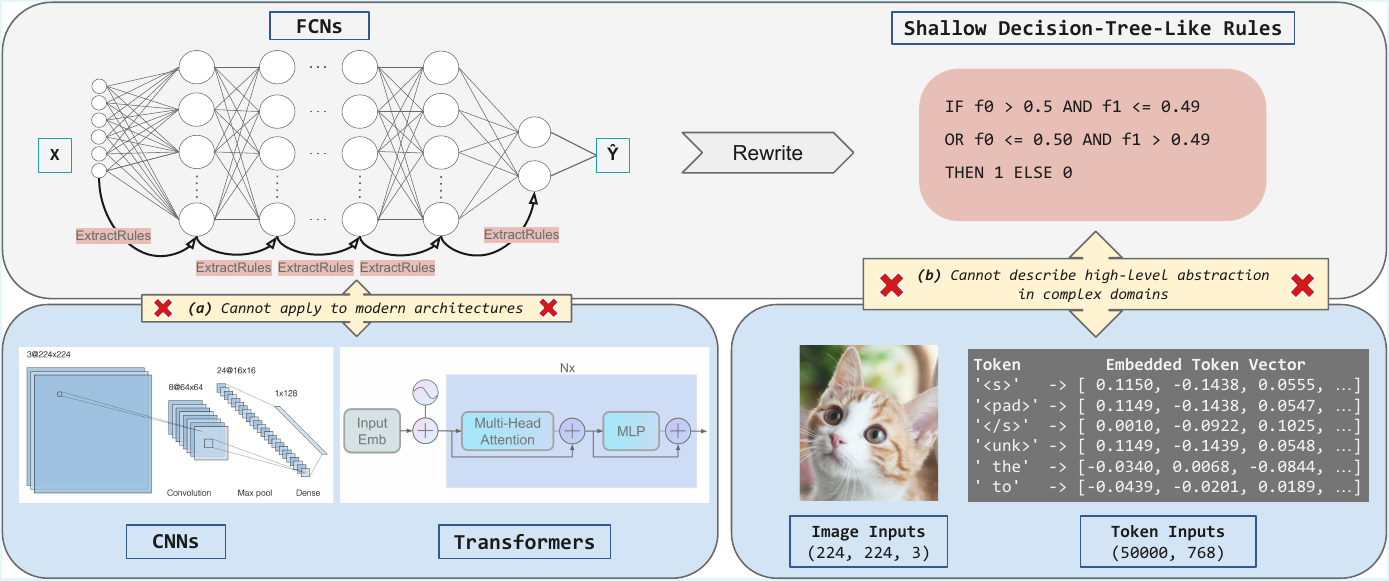}
    \caption{Existing rule-based methods fail to generalize to modern DNNs and their associated complex input domains.}
    \label{fig:intro}
\end{figure*}

Among various types of explanations for deep neural networks—such as attributions~\cite{grad_cam} and hidden semantics~\cite{BauZKO017}—rule-based methods that generate global logic rules over input sets, rather than local rules for individual samples, offer stronger interpretability and are highly preferred~\cite{pedreschi2019meobb}. However, most existing rule-based explanation methods \cite{RIPPER, DeepRED, ECLAIRE, CGXPLAIN} suffer from several limitations. We highlight three key issues, as illustrated in Figure~\ref{fig:intro}: (1) they mostly rely on layer-by-layer rule extraction and rewriting to derive final rules, which introduces scalability limitations; (2) they are primarily tailored to fully connected networks (FCNs) and fail to generalize to modern deep neural network (DNN) architectures such as convolutional neural networks and Transformers; (3) the rules they produce are often shallow and decision-tree-like, lacking the ability to capture high-level abstractions, which limits their effectiveness in complex domains.


To this end, we introduce \textsc{NeuroLogic}, a modern rule-based framework designed to address architectural dependence, limited scalability, and the shallow nature of existing decision rules. Our approach is inspired by Neural Activation Patterns (NAPs)~\cite{geng23,geng2024} which are subsets of neurons that consistently activate for inputs belonging to the same class. Specifically, for any given layer, we identify salient neurons for each class and determine their optimal activation thresholds, converting these neurons into \emph{hidden predicates}. These predicates represent high-level features learned by the model, where a true value indicates the presence of the corresponding feature in a given input. Based on these predicates, \textsc{NeuroLogic} constructs first-order logic (FOL) rules in a fully data-driven manner to approximate the internal behavior of neural networks.

The remaining challenge is to ground these hidden predicates in the original input space to ensure interpretability. Unlike existing approaches that can only produce shallow, decision-tree-like rules, \textsc{NeuroLogic} features a flexible design that supports a wide range of interpretable surrogate methods, such as program synthesis, to learn rules with richer and more expressive structures. It can also incorporate human prior knowledge as high-level abstractions of complex input domains to enable more efficient and meaningful grounding. To demonstrate its capabilities, we apply \textsc{NeuroLogic} to extract logic rules from Transformer-based sentiment analysis---a setting where traditional rule-extraction methods struggle to scale. \emph{To the best of our knowledge, this is the first approach capable of extracting global logic rules from modern, complex architectures such as Transformers.} We believe \textsc{NeuroLogic} represents a promising step toward opening the black box of deep neural networks. Our contributions are summarized as follows:



\begin{itemize}
\item We propose \textsc{NeuroLogic}, a novel framework for extracting interpretable global logic rules from deep neural networks. By abandoning the costly layer-wise rule extraction and substitution paradigm, \textsc{NeuroLogic} achieves greater scalability and broad architectural compatibility.
\item The decoupled design of \textsc{NeuroLogic} enables flexible grounding, allowing the generation of more abstract and interpretable rules that transcend the limitations of shallow, decision tree–based explanations.
\item Experimental results on small-scale benchmarks demonstrate that \textsc{NeuroLogic} produces more compact rules with higher efficiency than state-of-the-art methods, while maintaining strong fidelity and predictive accuracy.
\item We further showcase the practical feasibility of \textsc{NeuroLogic} in extracting meaningful logic rules and providing insights into the internal mechanisms of Transformers—an area where existing approaches struggle to scale effectively.
\end{itemize}

\section{Preliminary}
\label{sec:background}

\paragraph{Neural Networks for Classification Tasks} We consider a general deep neural network  \( N \) used for classification. Let \( z^{l}_i(x) \) denote the value of the \( i \)-th neuron at layer \( l \) for a given input \( x \). We do not assume any specific form for the transformation between layers, that is, the mapping from \( z^{l} \) to \( z^{l+1} \) can be arbitrary. This abstraction allows our analysis to be broadly applied across architectures.

The network \( N \) as a whole functions as 
\begin{align}
\mathbf{F}^{<N>}: X \to \mathbb{R}^{|C|},
\end{align}
mapping an input \( x \in X \) from the dataset to a score vector over the class set \( C \). The predicted class is then given by
\begin{align}
\hat{c} = \arg\max_{c \in C} \mathbf{F}^{<N>}_c(x).
\end{align}



\paragraph{First-Order Logic}  
First-Order Logic (FOL) is a formal language for stating interpretable rules about objects and their relations/attributes. It extends propositional logic by introducing \textit{quantifiers} such as:

\begin{itemize}
    \item \textit{Universal quantifier} (\( \forall \)): meaning “for all”, e.g., \(\forall x\, p(x)\) means \(p(x)\) holds for every \(x\).
    \item \textit{Existential quantifier} (\( \exists \)): meaning “there exists”, e.g., \(\exists x\, p(x)\) means there exists at least one \(x\) for which \(p(x)\) holds.
\end{itemize}

We focus on FOL rules in \textit{Disjunctive Normal Form (DNF)}, which are disjunctions (ORs) of conjunctions (ANDs) of \textit{predicates}.  

\begin{itemize}
    \item A \textit{predicate} is a simple condition or property on the input, e.g., \(p_i(x)\).  
    \item A \textit{clause} is a conjunction (AND) of predicates, such as \(p_1(x) \land p_2(x) \land \neg p_3(x)\).  
\end{itemize}

A DNF rule looks like a logical OR of multiple clauses:  
\begin{align}
\forall x, \quad \left(p_1(x) \land p_2(x)\right) \lor \left(p_3(x) \land \neg p_4(x)\right) \Rightarrow \textit{Label}(x) = c,
\end{align}
meaning that for every input \(x\), if any clause is satisfied, it is assigned to class \(c\). This structured form makes the rules easy to interpret and understand.

\section{The \textsc{NeuroLogic} Framework}
\label{sec:method}
\begin{figure*}[t] 
    \centering
    \includegraphics[width=\textwidth]{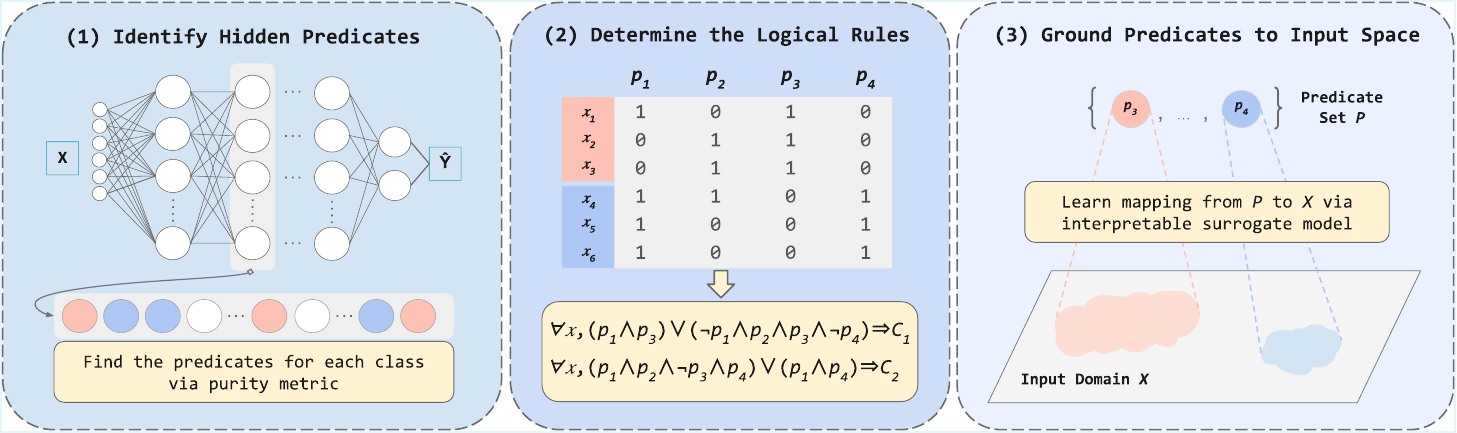}
    \caption{Overview of the \textsc{NeuroLogic} Framework.}
    \label{fig:overview}
\end{figure*}

In this section, we introduce \textsc{NeuroLogic}, a novel approach for extracting interpretable logic rules from DNNS. For clarity, we divide the \textsc{NeuroLogic} framework into three subtasks. An overview is illustrated in Figure~\ref{fig:overview}.

\subsection{Identifying Hidden Predicates}  
\label{sec:phase2}

For a given layer \( l \), we aim to identify a subset of neurons that are highly indicative of a particular class \( c \in C \). These neurons form what are known as Neural Activation Patterns (NAPs)~\cite{geng23,geng2024}. A neuron is considered part of the NAP for class \( c \) if its activation is consistently higher for inputs from class \( c \) compared to inputs from other classes. This behavior suggests that such neurons encode class-specific latent features at layer \( l \), as discussed in~\cite{geng2024}.

To identify the NAP for a specific class \( c \), we evaluate how selectively each neuron responds to class \( c \) versus other classes. Since each neuron’s activation is a scalar value, we can assess its discriminative power by learning a threshold \( t \). This threshold separates inputs from class \( c \) and those from other classes based on activation values.

Formally, we consider a neuron to support class \( c \) if its activation \( z_j^l(x) \) for input \( x \) satisfies \( z_j^l(x) \geq t \). If this condition holds, we classify \( x \) as belonging to class \( c \); otherwise, it is classified as not belonging to \( c \). To quantify the effectiveness of a threshold \( t \), we use the \textit{purity} metric, defined as:
\begin{align}
\text{Purity}(t) =\ & \frac{\left| \left\{ x \in X_c : z_j^{l}(x) \geq t \right\} \right|}{|X_c|} \nonumber \\
& + \frac{\left| \left\{ x \in X_{\neg c} : z_j^{l}(x) < t \right\} \right|}{|X_{\neg c}|}
\end{align}

Here, \( X_c \) denotes the set of inputs from class \( c \), while \( X_{\neg c} \) denotes inputs from all other classes. A high purity value means the neuron cleanly separates class \( c \) from others, whereas a low value suggests ambiguous or overlapping activation responses. We conduct a linear search to determine the optimal threshold \( t \) as its final purity.

In our implementation, for each neuron, we compute its purity with respect to each class to determine its class preference. Then, for each class, we rank the neurons by that purity and keep the top-$k$. These selected neurons are referred to as \textit{hidden predicates}, denoted as $P$, as they capture discriminative features that are highly specific to each class within the input space.

\subsection{Determining the Logical Rules}

Formally, a predicate \( p_j \) at layer \( l \), together with its corresponding threshold \( t_j \), is defined as \( p_j(x) := \mathbb{I}[ z_j^{(l)}(x) \geq t_j] \). In this context, a True (1) assignment indicates the presence of the specific latent feature of class \( c \) for input \( x \), while a False (0) assignment signifies its absence. Intuitively, the more predicates that fire, the stronger the evidence that $x$ belongs to class $c$. However, this raises the question: to what extent should we believe that \( x \) belongs to class \( c \) based on the pattern of predicate activations?

We address this question using a data-driven approach. 
Let $P_c^{(l)} = \{p_1, ..., p_m\}$ be the $m$ predicates retained for class $c.$ Evaluating $P_c^{(l)}$ on every class example $x \in X_c$ gives a multiset of binary vectors $p(x) \in \{0,1\}^m$. Each distinct vector can be treated as a clause, and the union of all clauses forms a DNF rule:
$$
\forall  x, \bigl( \bigvee_{v \in \mathcal{V}_c} \bigl(\bigwedge_{i: v_i=1} p_i(x) \wedge \bigwedge_{i: v_i = 0} \neg p_i(x) \bigr)\bigr) \implies Label(x) = c
$$
where $\mathcal{V}_c$ is the set of unique activation vectors for $X_c$ . For instance, suppose we have four predicates \( p_1(x), p_2(x), p_3(x), p_4(x) \) (we will omit \( x \) when the context is clear), and five distinct inputs yield the following patterns: \( (1, 1, 1, 1) \), \( (1, 1, 1, 0) \), \( (1, 1, 1, 0) \), \( (1, 1, 0, 1) \), and \( (1, 1, 0, 1) \). We can then construct a disjunctive normal form (DNF) expression to derive a rule:
\begin{align}
\forall x \quad (&(p_1 \land p_2 \land p_3 \land p_4) \lor (p_1 \land p_2 \land p_3 \land \neg p_4) \nonumber \\
&\lor (p_1 \land p_2 \land \neg p_3 \land p_4)) \Rightarrow \textit{Label}(x) = c.
\end{align}

In practice, these predicates behave as soft switches: their purity is imperfect, sometimes firing on inputs from $X_{\neg c}$. Consequently, the resulting DNF is best viewed as a comprehensive description and may include many predicates that are less relevant to the model’s actual classification behavior.

To address this, we apply a decision tree learner to \emph{distill} a more compact and representative version of the rule, which will serve as the (discriminative) rule-based model.


\subsection{Grounding Predicates to the Input Feature Space}
\label{sec:phase3}

The final step is to \emph{ground} these hidden predicates in the input space to make them human-interpretable. We adopt the definition of interpretability as the ability to explain model decisions in terms understandable to humans~\cite{doshi2017tarso}. Since "understandability"
is task and audience dependent, \textsc{NeuroLogic} is designed in a \emph{decoupled} fashion where any grounding method can be plugged in, allowing injection of domain knowledge. 

 This design also allows users to incorporate domain-specific knowledge where appropriate. Within the scope of this work, we present simple approaches for grounding predicates in simple input domains, as well as in the complex input domain of large vocabulary spaces for Transformers.

\emph{Exploring general grounding strategies for diverse tasks and models remains a challenge, and we believe it requires collective efforts from the whole research community.}

\paragraph{Grounding Predicates in Simple Input Domains}

For deep neural networks (DNNs) applied to tasks with simple input domains (e.g., tabular data), we aim to ground each predicate \( p_j \) directly in the raw input space. This enables more transparent and interpretable logic rules.

We reframe the grounding task as a supervised classification problem. For a given predicate \( p_j \), we collect input examples where the predicate is activated versus deactivated, and then learn a symbolic function that approximates this distinction.

Formally, for a target class \( c \) and predicate \( p_j \), we define the activation set and deactivation set, respectively, as
\begin{align}
D_1^{(j)} &= \{ x \in X_c \mid p_j(x) = 1 \}, \\
D_0^{(j)} &= \{ x \in X_c \mid p_j(x) = 0 \}.
\end{align}
These are combined into a labeled dataset
\begin{align}
D^{(j)} = \{ (x, y) \mid x \in D_1^{(j)} \cup D_0^{(j)},\ y = p_j(x) \}.
\end{align}

Then, to obtain expressive, compositional and human-readable logic rules as explanations, we employ program synthesis to learn a symbolic expression \( \phi_j \) from a domain-specific language (DSL) \( \mathcal{L} \). Unlike traditional decision-tree-like rules, the symbolic language \( \mathcal{L} \) is richer: a composable grammar over input features that supports not only logical and comparison operators but also linear combinations and nonlinear functions. Specifically, the language includes:

\begin{itemize}
    \item \textit{Atomic abstractions} formed by applying threshold comparisons to linear or nonlinear functions of the input features, for example,
    \begin{align}
    a := f(x) \leq \theta \quad \text{or} \quad f(x) > \theta,
    \end{align}
    where \( f(x) \) can be any linear or nonlinear transformation, such as polynomials, trigonometric functions, or other basis expansions.

  \item \textit{Logical operators} to combine these atomic abstractions into complex expressions:
  \begin{align}
  \phi ::= a \mid \neg \phi \mid \phi_1 \land \phi_2 \mid \phi_1 \lor \phi_2.
  \end{align}
\end{itemize}

The synthesis objective is to find an expression \( \phi_j \in \mathcal{L} \) that minimizes a combination of classification loss and complexity, formally:
\begin{align}
\phi_j \in \arg\min_{\phi \in \mathcal{L}} \left[ \mathcal{L}_{\mathrm{cls}}(\phi; D^{(j)}) + \lambda \cdot \Omega(\phi) \right],
\end{align}
where \( \mathcal{L}_{\mathrm{cls}} \) measures how well \( \phi \) approximates the predicate activations in \( D^{(j)} \), \( \Omega(\phi) \) penalizes the complexity of the expression (e.g., number of literals or tree depth), and \( \lambda \) balances the trade-off between accuracy and interpretability. 

This grounding approach also supports decision-tree-like rules, which are commonly used in existing methods. In this context, such rules can be viewed as a special case of the above atomic abstractions, where \( f(x) \) corresponds to individual features.

A simpler alternative is to leverage off-the-shelf decision tree algorithms: we train a decision tree classifier \( f_j^{\mathrm{DT}} \) such that
\begin{align}
f_j^{\mathrm{DT}}(x) \approx p_j(x), \quad \forall x \in X_c.
\end{align}
The resulting decision tree provides a simpler rule-based approximation of predicate activations, effectively grounding \( p_j \) in the input space in an interpretable manner.



\paragraph{Grounding predicates in the vocabulary space}

The input space in NLP domains (i.e., vocabulary spaces) is typically extremely large, making it difficult to ground rules onto raw feature vectors. In such domains, it is more effective to incorporate human prior knowledge like words, tokens, or linguistic structure that are more semantically meaningful and ultimately guide the predictions made by transformer-based models \cite{DBLP:conf/acl/TenneyDP19}. In light of this, we define a set of atomic abstractions over the vocabulary spaces. Each atomic abstraction corresponds to a template specifying keywords along with their associated lexical structures. To ground the learned hidden predicates to this domain knowledge, we leverage causal inference \cite{zarlenga2021efficient,DBLP:journals/corr/abs-2004-12265}.

Formally, let $\mathcal{A} = \{a_1, a_2, \dots, a_k\}$ be the set of atomic abstractions derived from domain knowledge (e.g., keywords or lexical patterns), and let \(p_j\) be a learned hidden predicate extracted from the model's internal representations, and \(x\) be an input instance (e.g., a text sample).

We define a causal intervention \(do(\neg a_i)\) as flipping the truth value of atomic abstraction \(a_i\) in the input \(x\) (e.g., masking the keyword associated with \(a_i\)). The grounding procedure tests whether flipping \(a_i\) changes the truth of the hidden predicate \(p_j\):
\begin{align}
\textit{If} \quad p_j(x) = \textit{True} \quad \textit{and} \quad p_j\bigl(do(\neg a_i)(x)\bigr) = \textit{False},
\end{align}
then we infer a causal dependence of \(p_j\) on \(a_i\), grounding \(p_j\) to the atomic abstraction \(a_i\).

By iterating over all atomic abstractions \(a_i \in \mathcal{A}\), we establish a mapping:
\begin{align}
G: p_j \mapsto \{a_i \in \mathcal{A} \mid \textit{flipping } a_i \textit{ negates } p_j \},
\end{align}
which grounds the hidden predicate \(p_j\) in terms of semantically meaningful domain knowledge.

\section{Evaluation}
\label{sec:eval}

\begin{table*}[h]
\centering
\renewcommand{\arraystretch}{1.2}
\resizebox{\textwidth}{!}{%
\begin{tabular}{c|l|c|c|c|c|c|}
\multicolumn{1}{c}{} & 
\multicolumn{1}{c}{\textbf{Method}} & 
\multicolumn{1}{c}{\textbf{Accuracy (\%)}} & 
\multicolumn{1}{c}{\textbf{Fidelity (\%)}} & 
\multicolumn{1}{c}{\textbf{Runtime (s)}} & 
\multicolumn{1}{c}{\textbf{Number of Clauses}} & 
\multicolumn{1}{c}{\textbf{Avg Clause Length}} \\        
\cline{2-7}
\multirow{5}{*}{\rotatebox{90}{XOR}} 
  & C5.0 &  52.6  $\pm$   0.2 &  53.0 $\pm$ 0.2 &   \textbf{0.1 $\pm$ 0.0} &   1 $\pm$ 0  &  1 $\pm$ 0   \\
  & ECLAIRE & 91.8 $\pm$ 1.0 & 91.4 $\pm$ 2.4 &  6.2 $\pm$ 0.4 & 87.0 $\pm$ 16.2 & 263.0 $\pm$  49.1
 \\
  & CGXPLAIN & \textbf{96.7 $\pm$ 1.7}  & \textbf{92.4 $\pm$ 1.1} & 9.1 $\pm$ 1.8  & \textbf{3.6 $\pm$ 1.8}  & 10.4 $\pm$ 7.2 \\
  & \textsc{NeuroLogic} & 89.6 $\pm$ 1.9 & 90.3 $\pm$ 1.6 &  1.2 $\pm$ 0.3 & 10.8 $\pm$ 3.5 & \textbf{6.8 $\pm$ 2.0} \\
\cline{2-7}
\multirow{5}{*}{\rotatebox{90}{MB-ER}} 
  & C5.0 &  92.7 $\pm$ 0.9  &  89.3 $\pm$ 1.0   &  20.3 $\pm$ 0.8 &  21.8 $\pm$ 3  &  72.4 $\pm$ 14.5  \\
  & ECLAIRE & \textbf{94.1 $\pm$ 1.6} & \textbf{94.7 $\pm$ 0.2} &   123.5 $\pm$ 36.8   &   48.3 $\pm$ 15.3 & 137.6 $\pm$ 24.7 \\
  & CGXPLAIN & 92.4 $\pm$ 0.7 & 94.7 $\pm$ 0.9 &  462.7 $\pm$ 34.0 & 5.9 $\pm$ 1.1 \ & 21.8 $\pm$ 3.4 \\
  & \textsc{NeuroLogic} & 92.8 $\pm$ 0.9 &  92.7 $\pm$ 1.4 & \textbf{6.0  $\pm$ 1.2} & \textbf{5.8  $\pm$ 1.0} & \textbf{3.7 $\pm$  0.2} \\
\cline{2-7}
\multirow{5}{*}{\rotatebox{90}{MB-HIST}} 
  & C5.0 &  87.9 $\pm$ 0.9  &  89.3 $\pm$ 1.0   &  16.06 $\pm$ 0.64  &  12.8 $\pm$ 3.1  &  35.2 $\pm$ 11.3 \\
  & ECLAIRE & 88.9 $\pm$ 2.3 & 89.4 $\pm$  1.8 & 174.5 $\pm$ 73.2 &  30.0 $\pm$  12.4 & 74.7 $\pm$  15.7 \\
  & CGXPLAIN &  89.4 $\pm$ 2.5 & 89.1 $\pm$ 3.6   & 285.3 $\pm$ 10.3 &  5.2 $\pm$ 1.9 & 27.8 $\pm$ 7.6 \\
  & \textsc{NeuroLogic} & \textbf{90.7 $\pm$  0.9} &  \textbf{92.0 $\pm$ 3.5}  & \textbf{2.3 $\pm$ 0.2}   & \textbf{3.6 $\pm$  1.6} & \textbf{2.7 $\pm$  0.3}\\
\cline{2-7}
\multirow{5}{*}{\rotatebox{90}{MAGIC}} 
  & C5.0 &  82.8 $\pm$  0.9  &  85.4 $\pm$  2.5  &  \textbf{1.9 $\pm$  0.1} &  57.8 $\pm$  4.5 &  208.7 $\pm$  37.6  \\
  & ECLAIRE & \textbf{84.6 $\pm$ 0.5} & 87.4 $\pm$ 1.2 &  240.0 $\pm$ 35.9  & 392.2 $\pm$ 73.9 & 1513.4 $\pm$  317.8 \\
  & CGXPLAIN & 84.4 $\pm$ 0.8 & \textbf{91.5 $\pm$ 1.3} &  44.6 $\pm$  2.9 & 7.4 $ \pm$  0.8 & 11.6 $\pm$ 1.9
\\
  & \textsc{NeuroLogic} &  \textbf{84.6 $\pm$ 0.5} & 90.8 $\pm$ 0.7 &  17.0  $\pm$ 1.5 & \textbf{6.0  $\pm$  0.0} & \textbf{3.6 $\pm$ 0.1} \\
\cline{2-7}
\end{tabular}
}
\caption{Comparison of rule-based explanation methods across different benchmarks. The best results are highlighted in bold.}
\label{tab:rule_comparison}
\end{table*}







In this section, we evaluate our approach, \textsc{NeuroLogic}, in two settings: (1) small-scale benchmarks and (2) transformer-based sentiment analysis. The former involves comparisons with baseline methods to assess \textsc{NeuroLogic} in terms of accuracy, efficiency, and interpretability. The latter focuses on a challenging, large-scale, real-world scenario where existing methods fail to scale, highlighting the practical viability and scalability of \textsc{NeuroLogic}.


\subsection{Small-Scale Benchmarks}

\paragraph{Setup and Baselines}
We evaluate \textsc{NeuroLogic} against popular rule-based explanation methods C5.0 \footnote{This represents the use of the C5.0 decision tree algorithm to learn rules in an end-to-end manner.}, ECLAIRE \cite{ECLAIRE}, and CGXPLAIN \cite{CGXPLAIN} on four standard interpretability benchmarks: XOR, MB-ER, MB-HIST, and MAGIC.  For each baseline, we use the original implementation and follow the authors’ recommended hyperparameters. We evaluate all methods using five metrics: accuracy, fidelity (agreement with the original model), runtime, number of clauses, and average clause length, to assess both interpretability and performance. Further details on the experimental setup are provided in the Appendix.

Notably, \textsc{NeuroLogic} consistently produces the most concise explanations, as reflected by both the number of clauses and the average clause length. In particular, it generates rule sets with substantially shorter average clause lengths—for example, on MB-HIST, it achieves $2.7 \pm 0.3$ compared to $27.8 \pm 7.6$ by the previous state-of-the-art, CGXPLAIN. This conciseness, along with fewer clauses, directly enhances interpretability and readability by reducing overall rule complexity. These results highlight a key advantage of \textsc{NeuroLogic} and align with our design goal of improving interpretability.

By avoiding the costly layer-wise rule extraction and substitution paradigm employed by ECLAIRE and CGXPLAIN, \textsc{NeuroLogic} achieves significantly higher efficiency. Although C5.0 can be faster in some cases by directly extracting rules from DNNs, it often suffers from lower fidelity, reduced accuracy, or the generation of overly complex rule sets. For example, while C5.0 can complete rule extraction on XOR in just 0.1 seconds, its accuracy is only around 52\%. In contrast, \textsc{NeuroLogic} consistently achieves strong performance in both fidelity and accuracy across all benchmarks. These results demonstrate that \textsc{NeuroLogic} strikes a favorable balance by effectively combining interpretability, computational efficiency, and faithfulness, outperforming existing rule-based methods.

\begin{figure*}[h]
    \centering
    \hspace*{\fill}%
    \begin{minipage}[t]{0.475\textwidth}
        \centering
        \includegraphics[width=\linewidth, keepaspectratio]{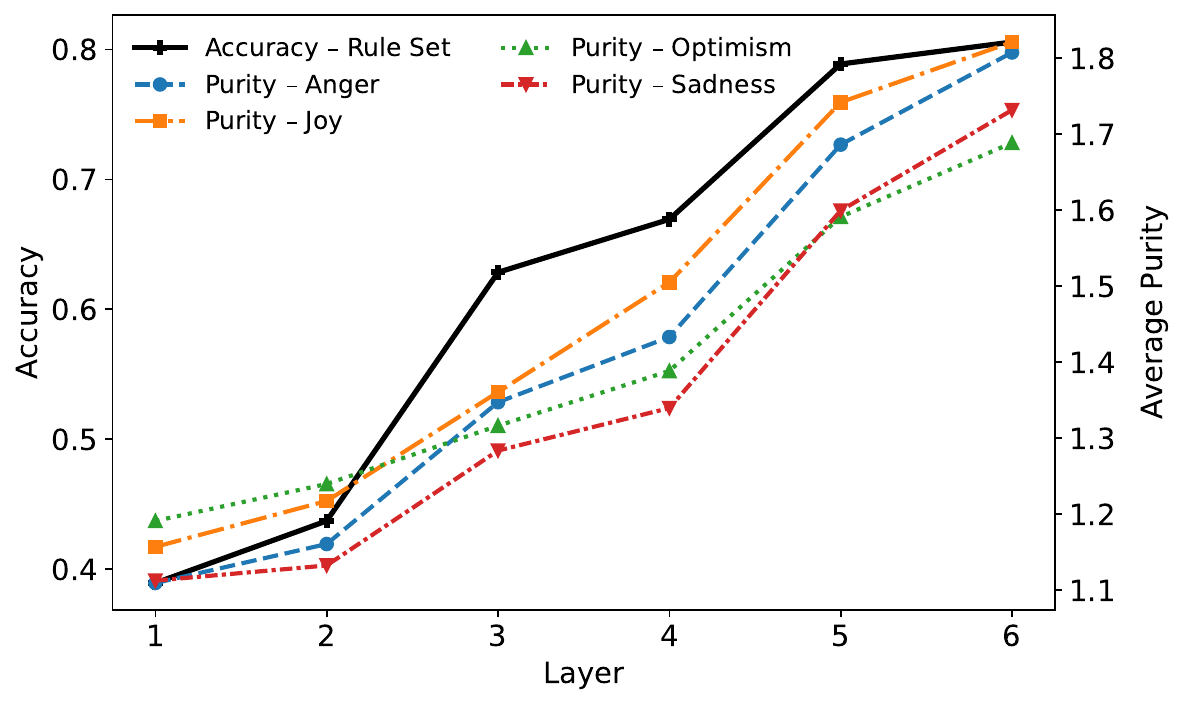}
        \captionof{figure}{The (average) purity of predicates correlates with the rule model accuracy as layers go deeper.}
        \label{fig:acc_avg_purity}
    \end{minipage}
    \hfill
    \begin{minipage}[t]{0.45\textwidth}
        \centering
        \includegraphics[width=\linewidth, keepaspectratio]{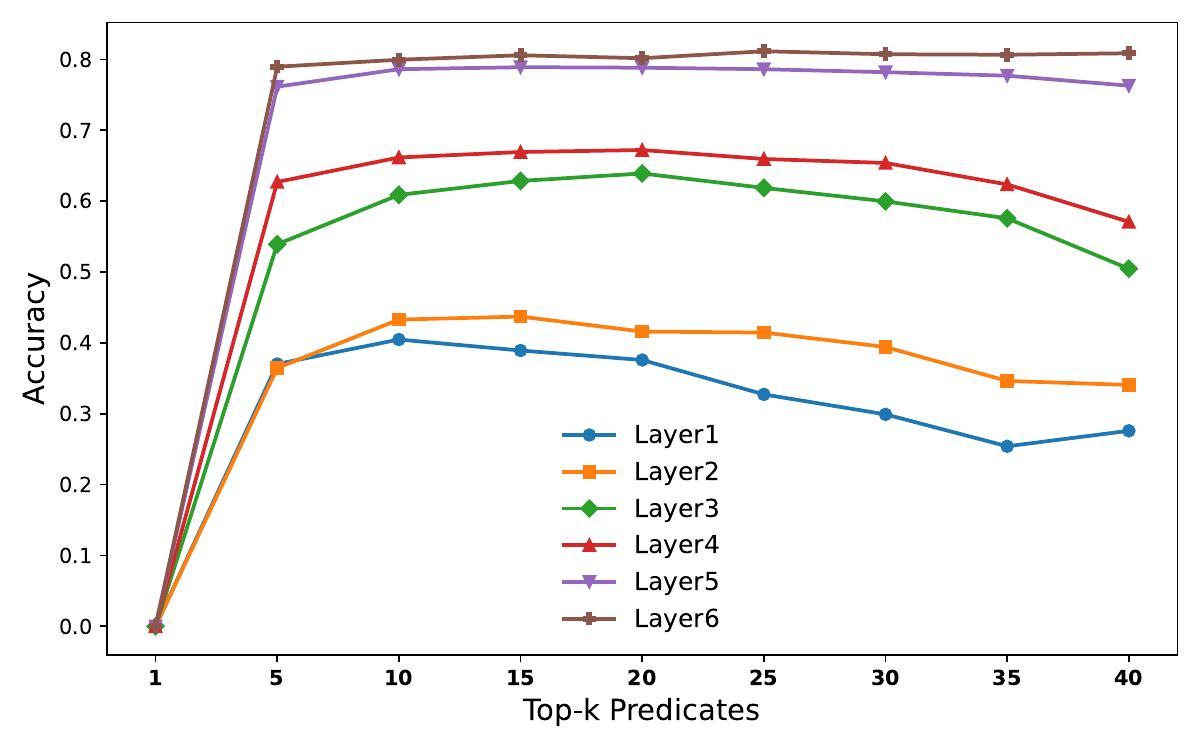}
        \captionof{figure}{The impact of the number of predicates affects the rule model.}
        \label{fig:topk_accuracy}
    \end{minipage}
    \hspace*{\fill}%
\end{figure*}

\begin{table*}[h]
    \centering
    \small
    \setlength{\tabcolsep}{3pt}
    \resizebox{\textwidth}{!}{%
    \begin{tabular}{c c c c c c c c c c c c c}
        \hline
        \multirow{2}{*}{\textbf{Class}}
            & \multicolumn{2}{c}{\textbf{Layer 1} (38.92\%)}
            & \multicolumn{2}{c}{\textbf{Layer 2} (43.70\%)}
            & \multicolumn{2}{c}{\textbf{Layer 3} (62.84\%)}
            & \multicolumn{2}{c}{\textbf{Layer 4} (66.92\%)}
            & \multicolumn{2}{c}{\textbf{Layer 5} (78.89\%)}
            & \multicolumn{2}{c}{\textbf{Layer 6} (80.58\%)}\\
        \cline{2-13}
            & \textit{\#\,Clauses} & \textit{Length}
            & \textit{\#\,Clauses} & \textit{Length}
            & \textit{\#\,Clauses} & \textit{Length}
            & \textit{\#\,Clauses} & \textit{Length}
            & \textit{\#\,Clauses} & \textit{Length}
            & \textit{\#\,Clauses} & \textit{Length}\\
        \hline
        Anger    & 70 & 4.29 & 91 & 4.27 & 91 & 4.82 & 75 & 4.51 & 42 & 4.19 & 33 & 4.15\\
        Joy      & 58 & 3.62 & 50 & 4.18 & 58 & 4.81 & 48 & 4.98 & 35 & 5.14 & 20 & 4.25\\
        Optimism & 34 & 4.88 & 32 & 4.38 & 47 & 5.60 & 49 & 5.65 & 26 & 4.65 & 23 & 5.57\\
        Sadness  & 78 & 4.76 & 53 & 4.36 & 84 & 5.25 & 73 & 3.78 & 46 & 4.72 & 38 & 4.92\\
        \hline
    \end{tabular}%
    }
    \caption{Number of clauses (\textit{\#\,Clauses}) and average clause length (\textit{Length}) for each emotion class across Transformer layers after pruning. Per-layer rule-set accuracy is shown in parentheses following the layer number.}
    \label{tab:rule_stats}
\end{table*}
\begin{table*}[t]
\centering
\small
\setlength{\tabcolsep}{3pt}
\resizebox{\textwidth}{!}{%
\begin{tabular}{|cc|cc|cc|cc|cc|cc|}
\hline
\multicolumn{2}{|c|}{\textbf{Layer 1}} &
\multicolumn{2}{c|}{\textbf{Layer 2}} &
\multicolumn{2}{c|}{\textbf{Layer 3}} &
\multicolumn{2}{c|}{\textbf{Layer 4}} &
\multicolumn{2}{c|}{\textbf{Layer 5}} &
\multicolumn{2}{c|}{\textbf{Layer 6}} \\
\cline{1-12}
Keyword & Pattern & Keyword & Pattern & Keyword & Pattern & Keyword & Pattern & Keyword & Pattern & Keyword & Pattern \\
\hline
the & at\_end & i & at\_start & it & after\_verb & sad & at\_end & sad & at\_end & sad & at\_end\\
i   & at\_end & i & before\_verb & you & after\_verb & in & after\_subject & sad & after\_verb & lost & after\_subject\\
of  & at\_end & user & at\_start & so & after\_subject & sad & after\_verb & depression & at\_end & depression & at\_end\\
when & at\_start & user & before\_subject & but & at\_start & sad & at\_start & me & after\_subject & sad & after\_verb\\
and & after\_subject & a & after\_subject & on & after\_verb & the & before\_verb & at & after\_verb & sad & after\_subject\\
at & after\_verb & i & after\_verb & a & at\_end & think & after\_subject & sad & after\_subject & sad & at\_start\\
be & after\_subject & a & after\_verb & of & at\_start & sad & after\_subject & sad & at\_start & sadness & at\_end\\
to & at\_end & is & after\_subject & you & after\_subject & depression & at\_end & sadness & at\_end & depression & at\_start\\
was & after\_verb & to & after\_verb & it & after\_subject & in & at\_start & depression & after\_verb & depressing & at\_end\\
sad & at\_end & i & after\_subject & just & at\_start & by & after\_verb & be & after\_verb & depressing & after\_subject\\
when & before\_subject & user & before\_verb & so & after\_verb & user & after\_verb & am & after\_subject & lost & at\_start\\
like & after\_subject & it & at\_start & can & after\_subject & be & after\_subject & was & before\_verb & nightmare & at\_end\\
when & before\_verb & is & at\_start & of & before\_verb & think & at\_start & at & after\_subject & sadness & after\_verb\\
like & after\_verb & and & after\_verb & can & before\_verb & with & after\_subject & at & at\_start & lost & at\_end\\
are & after\_subject & my & after\_verb & sad & after\_verb & really & after\_subject & depressing & at\_end & anxiety & after\_verb\\
\hline
\end{tabular}}
\caption{Top 15 keyword linguistic pattern pairs for class \emph{Sadness} learned by the top DNF rule across layers 1--6.}
\label{tab:top_15_patterns}
\end{table*}

\begin{table}[ht]
\centering
\renewcommand{\arraystretch}{1.2}   
\setlength{\tabcolsep}{8pt}         
\begin{tabular}{cll}
\hline
 & \textbf{EmoLex} & \textbf{NeuroLogic} \\ \hline
  & depression  & sad \\
  & bad        & depression \\
  & lost       & lost \\
  & terrorism  & depressing \\
 & sadness    & sadness \\
  & awful      & sadly \\
 & anxiety    & mourn \\
  & depressed  & nightmare \\
  & feeling    & anxiety \\
 & offended   & never \\ \hline
\textbf{F1} & 0.297 & \textbf{0.499} \\ \hline
\end{tabular}
\caption{Top-10 words for the \textit{Sadness} class from EmoLex and NeuroLogic. The bottom row reports the F1 scores.}
\label{tab:grounded_baselines}
\end{table}

\subsection{Transformer-based Sentiment Analysis}

\paragraph{Setup and Baselines}
We evaluate \textsc{NeuroLogic} on the \emph{Emotion} task from the \textsc{TweetEval} benchmark, which contains approximately 5,000 posts from Twitter, each labeled with one of four emotions: \textit{Anger}, \textit{Joy}, \textit{Optimism}, or \textit{Sadness} \cite{barbieri-etal-2020-tweeteval}. All experiments use the pretrained model, a 6-layer DistilBERT fine-tuned on the same TweetEval splits \cite{philschmid_distilbert_tweeteval_emotion_2024,sanh2020distilbertdistilledversionbert}. The pretrained model has a test accuracy of 80.59\%. The model contains approximately 66 million parameters, and we empirically validate that existing methods fail to efficiently scale to this level of complexity. For rule grounding, we approximate predicate-level interventions by masking tokens that instantiate an atomic abstract $a_i$ and flip an active DNF clause to \texttt{False}, thereby identifying $a_i$ as its causal grounder. In our study, each $a_i$ is defined as a \textit{(keyword, linguistic pattern)} pair, where the linguistic pattern may include structures such as \texttt{at\_start}. We benchmark the grounded rules produced by \textsc{NeuroLogic} against a classical purely lexical baseline.\footnote{To the best of our knowledge, no existing rule-extraction baseline is available for this task.} \textsc{EmoLex} \cite{mohammad13} tags a tweet as \emph{Sadness} whenever it contains any word from its emotion dictionary. This method relies on isolated keyword matching, with syntactical or other linguistic patterns ignored. Additional details are provided in the Appendix.






\paragraph{Identifying Predicates} We first extract hidden predicates from all six Transformer layers and observe that, as layers deepen, the predicates tend to exhibit higher purity, from average 1.1 to 1.8. This trend also correlates with the test accuracy from around 40 \% to 80\% of our rule-based model, as illustrated in Figure~\ref{fig:acc_avg_purity}. These results suggest that deeper layers capture more essential and task-relevant decision-making patterns, consistent with prior findings in \cite{geng23, geng2024}.  Another notable observation is that, surprisingly, a small number of predicates—specifically the top five—are often sufficient to explain the model’s behavior. As shown in Figure~\ref{fig:topk_accuracy}, including more predicates beyond this point can even reduce accuracy, particularly in shallower layers (Layers 1 and 2). Middle layers (Layers 3 and 4) are less affected, while deeper layers (Layers 5 and 6) remain relatively stable. Upon closer inspection, we find that this decline is due to the added predicates being noisier and less semantically meaningful, thereby introducing spurious patterns that degrade rule quality.

\paragraph{Constructing Rules}  Based on Figure~\ref{fig:topk_accuracy}, we select the top-15 predicates to construct the DNF rules, meaning that each clause initially consists of 15 predicates. However, after distillation, we find that, on average, fewer than five predicates are retained, as reported in Table~\ref{tab:rule_stats}. As a stand-alone classifier, the rule set distilled from \textit{Layer 6} achieves an accuracy of \emph{80.58\%}, on par with the neural model's accuracy ($80.59\%$).  Notably, the distilled DNF rule sets primarily consist of positive predicates, with negations rarely appearing. This indicates that the underlying neuron activations function more like \emph{selective filters}, each tuned to respond to specific input patterns rather than suppressing irrelevant ones. This aligns with the intuition that deeper transformer layers develop specialized units that favor and reinforce certain semantic or structural patterns, making the logic rules not only more compact but also more interpretable and faithful to the model's decision boundaries.

\paragraph{Grounding Rules} To simplify our analysis, we focus on the \emph{Sadness} class and the highest-scoring DNF rule per layer in Table \ref{tab:top_15_patterns}. We claim this is empirically justified: Figure \ref{fig:dnf-class-accuracy} (Appendix) shows that the class accuracy for each layer is explained significantly by the top DNF rule, so it effectively ``decides" whether an example is labelled \emph{Sadness} or not while the other rules handle outliers and more nuanced examples. In the earlier layers 1--2, high-frequency function keywords such as \textit{the}, \textit{i}, \textit{of}, and  \textit{at} mostly describe surface positions i.e \texttt{at\_end}. These words don't include any \emph{Sadness} emotional keywords but rather provides syntactic cues like subject boundaries and sentence structuring. This observation mirrors earlier probing attempts on Transformer layers \cite{tenney-etal-2019-bert, peters-etal-2018-dissecting}. In mid-layers 3--4, the introduction of explicit \emph{Sad} keywords (\textit{sad}, \textit{depression}) starts to mix in with anchors like \textit{in} and \textit{you}. This indicates a slow transition where emotional content is starting to get attended to, but overall linguistic patterns that encode local syntax are still required for rules to fire. Finally, in the deep layers 5--6, it is evident that the top rule fires nearly exclusively on keywords that convey \emph{Sadness} (\textit{sad}, \textit{lost}, \ textit {depression}, \textit{nightmare}, \textit{anxiety},  \textit{bad}). Each keyword appears numerous times paired with different linguistic patterns, with certain keywords being refined and pushed up (\textit{lost, sadness, sad,  depression}). Additionally, we also see a pattern collapse in later layers where many of the same keywords appear with multiple patterns. Together, these trends show that deeper predicates become less about local syntax and more about whether a salient semantic token is present anywhere in the input--an observation shared in many other findings \cite{de-vries-etal-2020-whats,  peters-etal-2018-dissecting}.

Table \ref{tab:grounded_baselines} compares the top-10 token cues for the class \emph{Sadness} extracted by each method. \textsc{NeuroLogic}'s top-10 list preserves core sadness cues like \textit{sad}, \textit{depression}, \textit{sadness}, \textit{depressing}, \textit{sadly}, \textit{depressed}, \textit{mourn}, \textit{anxiety} while promoting unique contextual hits like \textit{nightmare} and \textit{never} in place of more noisy terms like \textit{terrorism} or \textit{feeling}. Concretely, our method lifts the F1 from 0.297 to 0.499 by stripping out noisy cross-class terms without losing coverage.

\section{Related Work}
\label{sec:related}

Interpreting neural networks with logic rules has been explored since before the deep learning era. These approaches are typically categorized into two groups: pedagogical and decompositional methods~\cite{inter_survey, craven1994usaqt}. Pedagogical approaches approximate the network in an end-to-end manner. For example, classic decision tree algorithms such as CART~\cite{1984cart} and C4.5~\cite{1993cpfml} have been adapted to extract decision trees from trained neural networks~\cite{craven1996etsro, krishnan1999edtft, boz2002edtft}.  In contrast, decompositional methods leverage internal network information, such as structure and learned weights, to extract rules by analyzing the model's internal connections. A core challenge in rule extraction lies in identifying layerwise value ranges through these connections and mapping them back to input features. While recent works have explored more efficient search strategies~\cite{DeepRED, ECLAIRE, CGXPLAIN}, these methods typically scale only to very small networks due to the exponential growth of the search space with the number of attributes. Our proposed method, \textsc{NeuroLogic}, combines the efficiency of pedagogical approaches with the faithfulness of decompositional ones, making it scalable to modern DNN models. Its flexible design also enables the generation of more abstract and interpretable rules, moving beyond the limitations of shallow, decision tree–style explanations.

\section{Conclusion}
\label{sec:conclusion}

In this work, we introduce \textsc{NeuroLogic}, a novel framework for extracting interpretable logic rules from modern deep neural networks. \textsc{NeuroLogic} abandons the costly paradigm of layer-wise rule extraction and substitution, enabling greater scalability and architectural compatibility. Its decoupled design allows for flexible grounding, supporting the generation of more abstract and interpretable rules. We demonstrate the practical feasibility of \textsc{NeuroLogic} in extracting meaningful logic rules and providing deeper insights into the inner workings of Transformers.

\bibliography{main}

\begin{thebibliography}{34}
\providecommand{\natexlab}[1]{#1}

\bibitem[{Barbieri et~al.(2020)Barbieri, Camacho-Collados, Espinosa~Anke, and Neves}]{barbieri-etal-2020-tweeteval}
Barbieri, F.; Camacho-Collados, J.; Espinosa~Anke, L.; and Neves, L. 2020.
\newblock {T}weet{E}val: Unified Benchmark and Comparative Evaluation for Tweet Classification.
\newblock In Cohn, T.; He, Y.; and Liu, Y., eds., \emph{Findings of the Association for Computational Linguistics: EMNLP 2020}, 1644--1650. Online: Association for Computational Linguistics.

\bibitem[{Bau et~al.(2017)Bau, Zhou, Khosla, Oliva, and Torralba}]{BauZKO017}
Bau, D.; Zhou, B.; Khosla, A.; Oliva, A.; and Torralba, A. 2017.
\newblock Network Dissection: Quantifying Interpretability of Deep Visual Representations.
\newblock In \emph{2017 {IEEE} Conference on Computer Vision and Pattern Recognition, {CVPR} 2017, Honolulu, HI, USA, July 21-26, 2017}, 3319--3327. {IEEE} Computer Society.

\bibitem[{Bock et~al.(2004)Bock, Chilingarian, Gaug, Hakl, Hengstebeck, Ji{\v{r}}ina, Klaschka, Kotr{\v{c}}, Savick{\`y}, Towers et~al.}]{bock2004methods}
Bock, R.~K.; Chilingarian, A.; Gaug, M.; Hakl, F.; Hengstebeck, T.; Ji{\v{r}}ina, M.; Klaschka, J.; Kotr{\v{c}}, E.; Savick{\`y}, P.; Towers, S.; et~al. 2004.
\newblock Methods for multidimensional event classification: a case study using images from a Cherenkov gamma-ray telescope.
\newblock \emph{Nuclear Instruments and Methods in Physics Research Section A: Accelerators, Spectrometers, Detectors and Associated Equipment}, 516(2-3): 511--528.

\bibitem[{Boz(2002)}]{boz2002edtft}
Boz, O. 2002.
\newblock Extracting decision trees from trained neural networks.
\newblock In \emph{Proceedings of the Eighth {ACM} {SIGKDD} International Conference on Knowledge Discovery and Data Mining, July 23-26, 2002, Edmonton, Alberta, Canada}, 456--461. {ACM}.

\bibitem[{Breiman et~al.(1984)Breiman, Friedman, Olshen, and Stone}]{1984cart}
Breiman, L.; Friedman, J.~H.; Olshen, R.~A.; and Stone, C.~J. 1984.
\newblock \emph{Classification and Regression Trees}.
\newblock Wadsworth.
\newblock ISBN 0-534-98053-8.

\bibitem[{Cohen(1995)}]{RIPPER}
Cohen, W.~W. 1995.
\newblock Fast Effective Rule Induction.
\newblock In Prieditis, A.; and Russell, S., eds., \emph{Machine Learning, Proceedings of the Twelfth International Conference on Machine Learning, Tahoe City, California, USA, July 9-12, 1995}, 115--123. Morgan Kaufmann.

\bibitem[{Craven and Shavlik(1994)}]{craven1994usaqt}
Craven, M.~W.; and Shavlik, J.~W. 1994.
\newblock Using Sampling and Queries to Extract Rules from Trained Neural Networks.
\newblock In Cohen, W.~W.; and Hirsh, H., eds., \emph{Machine Learning, Proceedings of the Eleventh International Conference, Rutgers University, New Brunswick, NJ, USA, July 10-13, 1994}, 37--45. Morgan Kaufmann.

\bibitem[{Craven and Shavlik(1995)}]{craven1996etsro}
Craven, M.~W.; and Shavlik, J.~W. 1995.
\newblock Extracting Tree-Structured Representations of Trained Networks.
\newblock In Touretzky, D.~S.; Mozer, M.; and Hasselmo, M.~E., eds., \emph{Advances in Neural Information Processing Systems 8, NIPS, Denver, CO, USA, November 27-30, 1995}, 24--30. {MIT} Press.

\bibitem[{de~Vries, van Cranenburgh, and Nissim(2020)}]{de-vries-etal-2020-whats}
de~Vries, W.; van Cranenburgh, A.; and Nissim, M. 2020.
\newblock What{'}s so special about {BERT}{'}s layers? A closer look at the {NLP} pipeline in monolingual and multilingual models.
\newblock In Cohn, T.; He, Y.; and Liu, Y., eds., \emph{Findings of the Association for Computational Linguistics: EMNLP 2020}, 4339--4350. Online: Association for Computational Linguistics.

\bibitem[{Doshi-Velez and Kim(2017)}]{doshi2017tarso}
Doshi-Velez, F.; and Kim, B. 2017.
\newblock Towards A Rigorous Science of Interpretable Machine Learning.
\newblock \emph{arXiv: Machine Learning}.

\bibitem[{Geng et~al.(2023)Geng, Le, Xu, Wang, Gurfinkel, and Si}]{geng23}
Geng, C.; Le, N.; Xu, X.; Wang, Z.; Gurfinkel, A.; and Si, X. 2023.
\newblock Towards Reliable Neural Specifications.
\newblock In Krause, A.; Brunskill, E.; Cho, K.; Engelhardt, B.; Sabato, S.; and Scarlett, J., eds., \emph{International Conference on Machine Learning, {ICML} 2023, 23-29 July 2023, Honolulu, Hawaii, {USA}}, volume 202 of \emph{Proceedings of Machine Learning Research}, 11196--11212. {PMLR}.

\bibitem[{Geng et~al.(2024)Geng, Wang, Ye, Liao, and Si}]{geng2024}
Geng, C.; Wang, Z.; Ye, H.; Liao, S.; and Si, X. 2024.
\newblock Learning Minimal NAP Specifications for Neural Network Verification.
\newblock \emph{arXiv preprint arXiv:2404.04662}.

\bibitem[{Guidotti et~al.(2018)Guidotti, Monreale, Turini, Pedreschi, and Giannotti}]{guidotti2018asomf}
Guidotti, R.; Monreale, A.; Turini, F.; Pedreschi, D.; and Giannotti, F. 2018.
\newblock A Survey Of Methods For Explaining Black Box Models.
\newblock \emph{CoRR}, abs/1802.01933.

\bibitem[{He et~al.(2016)He, Zhang, Ren, and Sun}]{2016drlfi}
He, K.; Zhang, X.; Ren, S.; and Sun, J. 2016.
\newblock Deep Residual Learning for Image Recognition.
\newblock In \emph{2016 {IEEE} Conference on Computer Vision and Pattern Recognition, {CVPR} 2016, Las Vegas, NV, USA, June 27-30, 2016}, 770--778. {IEEE} Computer Society.

\bibitem[{Hemker, Shams, and Jamnik(2023)}]{CGXPLAIN}
Hemker, K.; Shams, Z.; and Jamnik, M. 2023.
\newblock CGXplain: Rule-Based Deep Neural Network Explanations Using Dual Linear Programs.
\newblock In Chen, H.; and Luo, L., eds., \emph{Trustworthy Machine Learning for Healthcare - First International Workshop, {TML4H} 2023, Virtual Event, May 4, 2023, Proceedings}, volume 13932 of \emph{Lecture Notes in Computer Science}, 60--72. Springer.

\bibitem[{Krishnan, Sivakumar, and Bhattacharya(1999)}]{krishnan1999edtft}
Krishnan, R.; Sivakumar, G.; and Bhattacharya, P. 1999.
\newblock Extracting decision trees from trained neural networks.
\newblock \emph{Pattern Recognit.}, 32(12): 1999--2009.

\bibitem[{Krizhevsky, Sutskever, and Hinton(2012)}]{2012icwdc}
Krizhevsky, A.; Sutskever, I.; and Hinton, G. 2012.
\newblock ImageNet Classification with Deep Convolutional Neural Networks.
\newblock In Bartlett, P.~L.; Pereira, F. C.~N.; Burges, C. J.~C.; Bottou, L.; and Weinberger, K.~Q., eds., \emph{Advances in Neural Information Processing Systems 25: 26th Annual Conference on Neural Information Processing Systems 2012. Proceedings of a meeting held December 3-6, 2012, Lake Tahoe, Nevada, United States}, 1106--1114.

\bibitem[{Lipton(2016)}]{lipton2016tmomi}
Lipton, Z.~C. 2016.
\newblock The Mythos of Model Interpretability.
\newblock \emph{CoRR}, abs/1606.03490.

\bibitem[{Mohammad and Turney(2013)}]{mohammad13}
Mohammad, S.~M.; and Turney, P.~D. 2013.
\newblock CROWDSOURCING A WORD–EMOTION ASSOCIATION LEXICON.
\newblock \emph{Computational Intelligence}, 29(3): 436--465.

\bibitem[{Pedreschi et~al.(2019)Pedreschi, Giannotti, Guidotti, Monreale, Ruggieri, and Turini}]{pedreschi2019meobb}
Pedreschi, D.; Giannotti, F.; Guidotti, R.; Monreale, A.; Ruggieri, S.; and Turini, F. 2019.
\newblock Meaningful Explanations of Black Box {AI} Decision Systems.
\newblock In \emph{The Thirty-Third {AAAI} Conference on Artificial Intelligence, {AAAI} 2019, The Thirty-First Innovative Applications of Artificial Intelligence Conference, {IAAI} 2019, The Ninth {AAAI} Symposium on Educational Advances in Artificial Intelligence, {EAAI} 2019, Honolulu, Hawaii, USA, January 27 - February 1, 2019}, 9780--9784. {AAAI} Press.

\bibitem[{Pereira et~al.(2016)Pereira, Chin, Rueda, Vollan, Provenzano, Bardwell, Pugh, Jones, Russell, Sammut et~al.}]{pereira2016somatic}
Pereira, B.; Chin, S.-F.; Rueda, O.~M.; Vollan, H.-K.~M.; Provenzano, E.; Bardwell, H.~A.; Pugh, M.; Jones, L.; Russell, R.; Sammut, S.-J.; et~al. 2016.
\newblock The somatic mutation profiles of 2,433 breast cancers refine their genomic and transcriptomic landscapes.
\newblock \emph{Nature communications}, 7(1): 11479.

\bibitem[{Peters et~al.(2018)Peters, Neumann, Zettlemoyer, and Yih}]{peters-etal-2018-dissecting}
Peters, M.~E.; Neumann, M.; Zettlemoyer, L.; and Yih, W.-t. 2018.
\newblock Dissecting Contextual Word Embeddings: Architecture and Representation.
\newblock In Riloff, E.; Chiang, D.; Hockenmaier, J.; and Tsujii, J., eds., \emph{Proceedings of the 2018 Conference on Empirical Methods in Natural Language Processing}, 1499--1509. Brussels, Belgium: Association for Computational Linguistics.

\bibitem[{Quinlan(1993)}]{1993cpfml}
Quinlan, J.~R. 1993.
\newblock \emph{{C4.5:} Programs for Machine Learning}.
\newblock Morgan Kaufmann.
\newblock ISBN 1-55860-238-0.

\bibitem[{Sanh et~al.(2020)Sanh, Debut, Chaumond, and Wolf}]{sanh2020distilbertdistilledversionbert}
Sanh, V.; Debut, L.; Chaumond, J.; and Wolf, T. 2020.
\newblock DistilBERT, a distilled version of BERT: smaller, faster, cheaper and lighter.
\newblock arXiv:1910.01108.

\bibitem[{Schmid(2024)}]{philschmid_distilbert_tweeteval_emotion_2024}
Schmid, P. 2024.
\newblock {philschmid/DistilBERT-tweet-eval-emotion}.
\newblock \url{https://huggingface.co/philschmid/DistilBERT-tweet-eval-emotion}.
\newblock Hugging Face model card, version accessed 31 Jul 2025.

\bibitem[{Selvaraju et~al.(2017)Selvaraju, Das, Vedantam, Cogswell, Parikh, and Batra}]{grad_cam}
Selvaraju, R.~R.; Das, A.; Vedantam, R.; Cogswell, M.; Parikh, D.; and Batra, D. 2017.
\newblock Grad-CAM: Visual Explanations from Deep Networks via Gradient-Based Localization.
\newblock In \emph{2017 IEEE International Conference on Computer Vision (ICCV)}, 618--626.

\bibitem[{Sutskever, Vinyals, and Le(2014)}]{2014stslw}
Sutskever, I.; Vinyals, O.; and Le, Q.~V. 2014.
\newblock Sequence to Sequence Learning with Neural Networks.
\newblock In Ghahramani, Z.; Welling, M.; Cortes, C.; Lawrence, N.~D.; and Weinberger, K.~Q., eds., \emph{Advances in Neural Information Processing Systems 27: Annual Conference on Neural Information Processing Systems 2014, December 8-13 2014, Montreal, Quebec, Canada}, 3104--3112.

\bibitem[{Tenney, Das, and Pavlick(2019{\natexlab{a}})}]{DBLP:conf/acl/TenneyDP19}
Tenney, I.; Das, D.; and Pavlick, E. 2019{\natexlab{a}}.
\newblock {BERT} Rediscovers the Classical {NLP} Pipeline.
\newblock In Korhonen, A.; Traum, D.~R.; and M{\`{a}}rquez, L., eds., \emph{Proceedings of the 57th Conference of the Association for Computational Linguistics, {ACL} 2019, Florence, Italy, July 28- August 2, 2019, Volume 1: Long Papers}, 4593--4601. Association for Computational Linguistics.

\bibitem[{Tenney, Das, and Pavlick(2019{\natexlab{b}})}]{tenney-etal-2019-bert}
Tenney, I.; Das, D.; and Pavlick, E. 2019{\natexlab{b}}.
\newblock {BERT} Rediscovers the Classical {NLP} Pipeline.
\newblock In Korhonen, A.; Traum, D.; and M{\`a}rquez, L., eds., \emph{Proceedings of the 57th Annual Meeting of the Association for Computational Linguistics}, 4593--4601. Florence, Italy: Association for Computational Linguistics.

\bibitem[{Vig et~al.(2020)Vig, Gehrmann, Belinkov, Qian, Nevo, Singer, and Shieber}]{DBLP:journals/corr/abs-2004-12265}
Vig, J.; Gehrmann, S.; Belinkov, Y.; Qian, S.; Nevo, D.; Singer, Y.; and Shieber, S.~M. 2020.
\newblock Causal Mediation Analysis for Interpreting Neural {NLP:} The Case of Gender Bias.
\newblock \emph{CoRR}, abs/2004.12265.

\bibitem[{Zarlenga, Shams, and Jamnik(2021{\natexlab{a}})}]{ECLAIRE}
Zarlenga, M.~E.; Shams, Z.; and Jamnik, M. 2021{\natexlab{a}}.
\newblock Efficient Decompositional Rule Extraction for Deep Neural Networks.
\newblock \emph{CoRR}, abs/2111.12628.

\bibitem[{Zarlenga, Shams, and Jamnik(2021{\natexlab{b}})}]{zarlenga2021efficient}
Zarlenga, M.~E.; Shams, Z.; and Jamnik, M. 2021{\natexlab{b}}.
\newblock Efficient decompositional rule extraction for deep neural networks.
\newblock \emph{arXiv preprint arXiv:2111.12628}.

\bibitem[{Zhang et~al.(2021)Zhang, Ti{\~{n}}o, Leonardis, and Tang}]{inter_survey}
Zhang, Y.; Ti{\~{n}}o, P.; Leonardis, A.; and Tang, K. 2021.
\newblock A Survey on Neural Network Interpretability.
\newblock \emph{{IEEE} Trans. Emerg. Top. Comput. Intell.}, 5(5): 726--742.

\bibitem[{Zilke, Loza~Menc{\'\i}a, and Janssen(2016)}]{DeepRED}
Zilke, J.~R.; Loza~Menc{\'\i}a, E.; and Janssen, F. 2016.
\newblock Deepred--rule extraction from deep neural networks.
\newblock In \emph{Discovery Science: 19th International Conference, DS 2016, Bari, Italy, October 19--21, 2016, Proceedings 19}, 457--473. Springer.

\end{thebibliography}



\clearpage

\UseRawInputEncoding

\appendix

\section{Additional Details on Small-Scale Benchmarks}


All experiments were conducted on a desktop equipped with a 2GHz Intel i7 processor and 32 GB of RAM. For each baseline, we used the original implementation and followed the authors’ recommended hyperparameters to ensure a fair comparison. We performed all experiments across five different random folds to initialize the train-test splits, the random initialization of the DNN, and the random inputs for the baselines. Regarding the metric of average clause length, there appears to be a discrepancy in how it is computed in \cite{ECLAIRE} and \cite{CGXPLAIN}. Specifically, \cite{ECLAIRE} seems to underestimate the average clause length. To ensure consistency and accuracy, we adopt the computation method used in \cite{CGXPLAIN}.

To maintain consistency, we used the same DNN topology (i.e., number and depth of layers) as in the experiments reported by \cite{ECLAIRE}. For \textsc{NeuroLogic}, we applied it to the last hidden layer and used the C5.0 decision tree as the grounding method for optimal efficiency. Below is a detailed description of each dataset:

\paragraph{MAGIC.} The MAGIC dataset simulates the detection of high-energy gamma particles versus background cosmic hadrons using imaging signals captured by a ground-based atmospheric Cherenkov telescope \citep{bock2004methods}. It consists of 19,020 samples with 10 handcrafted features extracted from the telescope’s “shower images.” The dataset is moderately imbalanced, with approximately 35\% of instances belonging to the minority (gamma) class.

\paragraph{Metabric-ER.} This biomedical dataset is constructed from the METABRIC cohort and focuses on predicting Estrogen Receptor (ER) status---a key immunohistochemical marker for breast cancer---based on 1,000 features, including tumor characteristics, gene expression levels, clinical variables, and survival indicators. Of the 1,980 patients, roughly 24\% are ER-positive, indicating the presence of hormone receptors that influence tumor growth.

\paragraph{Metabric-Hist.} Also derived from the METABRIC cohort \citep{pereira2016somatic}, this dataset uses the mRNA expression profiles of 1,694 patients (spanning 1,004 genes) to classify tumors into two major histological subtypes: Invasive Lobular Carcinoma (ILC) and Invasive Ductal Carcinoma (IDC). Positive diagnoses (ILC) account for only 8.7\% of all samples, resulting in a highly imbalanced classification setting.

\paragraph{XOR.} A synthetic dataset commonly used as a benchmark for rule-based models. Each instance \( \mathbf{x}^{(i)} \in [0, 1]^{10} \) is sampled independently from a uniform distribution. Labels are assigned according to a non-linear XOR logic over the first two dimensions:
\[
y^{(i)} = \text{round}(x^{(i)}_1) \oplus \text{round}(x^{(i)}_2),
\]
where \( \oplus \) denotes the logical XOR operation. The dataset contains 1,000 instances.


\section{Additional Details on Transformer-Based Sentiment Analysis}

All experiments are conducted on a machine running Ubuntu 22.04 LTS, equipped with an NVIDIA A100 GPU (40 GB VRAM), 85 GB of RAM, and dual Intel Xeon CPUs.

\paragraph{EmoLex.} We use the NRC Word-Emotion Association Lexicon \cite{mohammad13}. Tweets are lower-cased and split into alphabetical word fragments with regex. The tweet is assigned emotion $e$ iff any word appears in the EmoLex list for $e$. No lemmatisation, emoji handling, or other heuristics are applied. 


\subsection{Grounding Rule Templates Procedure}
Given DNFs extracted in \S\ref{sec:phase2}, we ground each DNF to lexical templates of the underlying text. Our implementation (\texttt{causal\_word\_lexical\_batched} does the following:
\paragraph{Implementation}
We use \textbf{spaCy 3.7} (\texttt{en\_core\_web\_sm}) for sentence segmentation, POS tags, and dependency arcs. 1) \textbf{Casual test.} For every neuron-predicate in the learned DNF, we \emph{mask} one candidate word. If the forward pass flips the DNF class prediction i.e any predicate in the DNF flips, the word is deemed causal. We then fit this word into the possible templates.
\\
2) \textbf{Template types.} 
Once a word is deemed \emph{causal}, we map it to the first matching
template in the following order:

\begin{enumerate}
  \item \textbf{\textsc{is\_hashtag}}: word starts with ``\#''.
  \item \textbf{\textsc{at\_start} / \textsc{at\_end}}: word’s index
        within its sentence falls in the first or last 20 \% of tokens
        (\(\alpha = 0.20\)).
  \item \textbf{\textsc{before/after\_subject}}: using
        \texttt{spaCy}, locate the first \texttt{nsubj/nsubjpass}; the
        word is \textsc{before} or \textsc{after} if it appears within a
        $\pm$ 6-token window of that subject.
  \item \textbf{\textsc{before/after\_verb}}: same window logic around
        the first main \texttt{VERB}.
  \item \textbf{\textsc{exists}}: general template, applied to all templates.
\end{enumerate}
This assignment yields the (\textit{word}, \textit{template}) pair that forms the grounded rules. \\
3) \textbf{Scoring \& ordering.}
For every (\textit{word},\,\textit{template}) rule, we compute a \emph{support score}
\[
s = \operatorname{idf}(w)\,
    \frac{\texttt{flips}(w,t)}{\texttt{total}(w,t)},
\quad
\operatorname{idf}(w)=\log\!\frac{N_{\text{docs}}+1}{\text{df}(w)+1}.
\]
Templates with \(s\ge\tau\) (\(\tau=0.03\)) are kept. The final rule list
for each class is sorted in \textbf{descending \(s\)} so highest score appears first.

\subsection{Top DNF rule accuracy for each class}

\begin{figure}[h]
\centering
\includegraphics[width=0.45\textwidth]{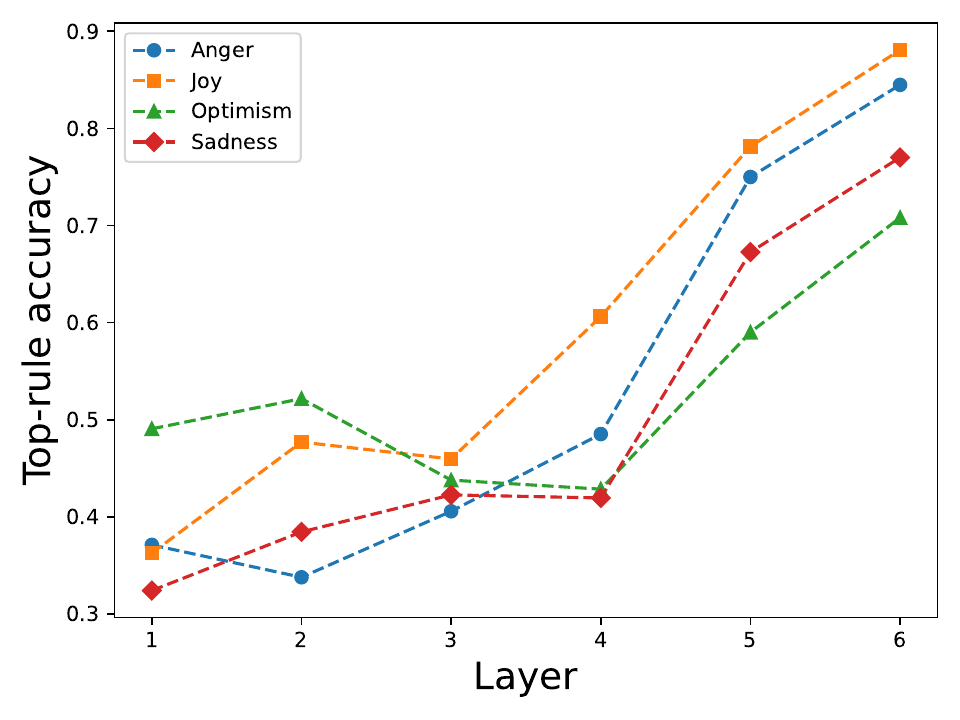}
\caption{Top DNF rule accuracy for each class by layer.}
\label{fig:dnf-class-accuracy}
\end{figure}

We report the class-wise accuracy achieved by the top DNF rule at each layer in Figure~\ref{fig:dnf-class-accuracy}. The results show that each layer's behavior can be effectively and consistently explained by its corresponding top DNF rule, demonstrating a strong alignment between the rule and the model's internal representations.

\newpage





\subsection{Code}
All code used for our experiments is available in the following GitHub repository: \texttt{github.com/NeuroLogic2026/NeuroLogic}. 

The sample code for purity-based predicates extraction.
\begin{lstlisting}[language=Python,label={lst:purity_rules}]
import torch
from typing import Dict, List, Tuple

def purity_rules(
    z_cls: torch.Tensor,          # (N, H) CLS activations
    y: torch.Tensor,              # (N,) integer class labels
    k: int = 15                   # top-k neurons per class
) -> Tuple[
        Dict[int, List[Tuple[int, float, int]]],   # rules[c] = [(neuron, τ, support)]
        Tuple[int, int, float, float, int]         # best (class, neuron, τ, purity, support)
]:
    num_samples, hidden_size = z_cls.shape
    num_classes = int(y.max().item()) + 1

    purity   = torch.empty(num_classes, hidden_size)
    thr_mat  = torch.empty(num_classes, hidden_size)
    supp_mat = torch.empty(num_classes, hidden_size, dtype=torch.long)

    class_counts = torch.bincount(y, minlength=num_classes)

    for j in range(hidden_size):
        a = z_cls[:, j]
        idx = torch.argsort(a, descending=True)
        a_sorted, y_sorted = a[idx], y[idx]

        one_hot = torch.nn.functional.one_hot(
            y_sorted, num_classes=num_classes
        ).cumsum(0)
        total_seen = torch.arange(1, num_samples + 1)

        for c in range(num_classes):
            tp = one_hot[:, c]
            fp = total_seen - tp
            tn = (num_samples - class_counts[c]) - fp

            tp_rate = tp.float() / class_counts[c].clamp_min(1)
            tn_rate = tn.float() / (num_samples - class_counts[c]).clamp_min(1)
            p_scores = tp_rate + tn_rate

            best = torch.argmax(p_scores)
            purity[c, j]  = p_scores[best]
            thr_mat[c, j] = a_sorted[best].item()
            supp_mat[c, j] = total_seen[best].item()

    rules: Dict[int, List[Tuple[int, float, int]]] = {
        c: [
            (j, thr_mat[c, j].item(), supp_mat[c, j].item())
            for j in torch.topk(purity[c], k=min(k, hidden_size)).indices.tolist()
        ]
        for c in range(num_classes)
    }

    best_c, best_j = divmod(purity.argmax().item(), hidden_size)
    best_neuron = (
        best_c,
        best_j,
        thr_mat[best_c, best_j].item(),
        purity[best_c, best_j].item(),
        supp_mat[best_c, best_j].item(),
    )
    return rules, best_neuron
\end{lstlisting}

\subsection{Token Position Analysis}
\begin{figure*}[!t]
\centering
\includegraphics[width=\textwidth]{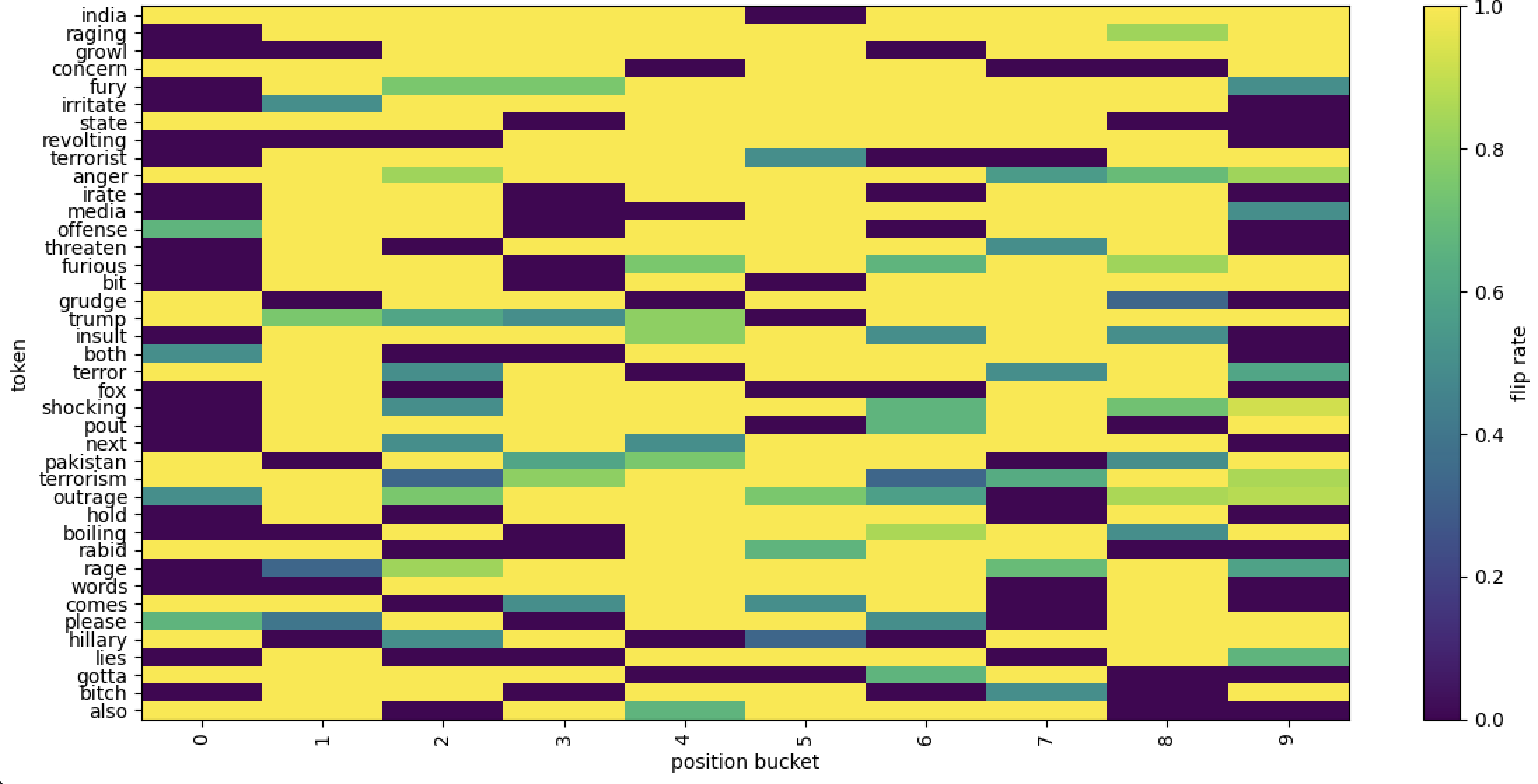}
\caption{Heat map of keywords by positional bucket for class \emph{Anger}.}
\label{fig:anger-heatmap}
\end{figure*}

\begin{figure*}[h]
\centering
\includegraphics[width=\textwidth]{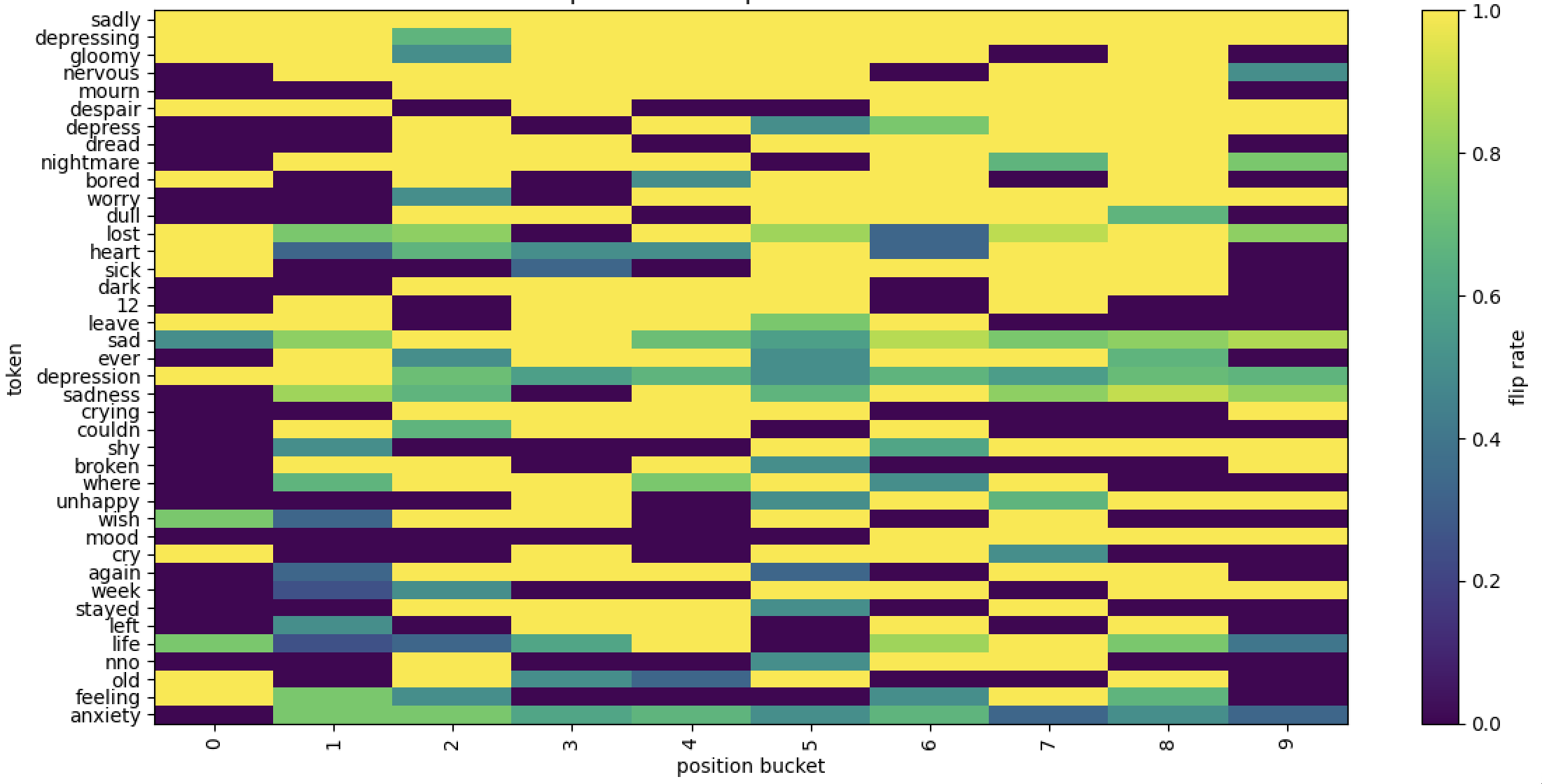}
\caption{Heat map of keywords by positional bucket for class \emph{Sadness}.}
\label{fig:sadness-heatmap}
\end{figure*}

\begin{figure*}[t]
\centering
\includegraphics[width=\textwidth]{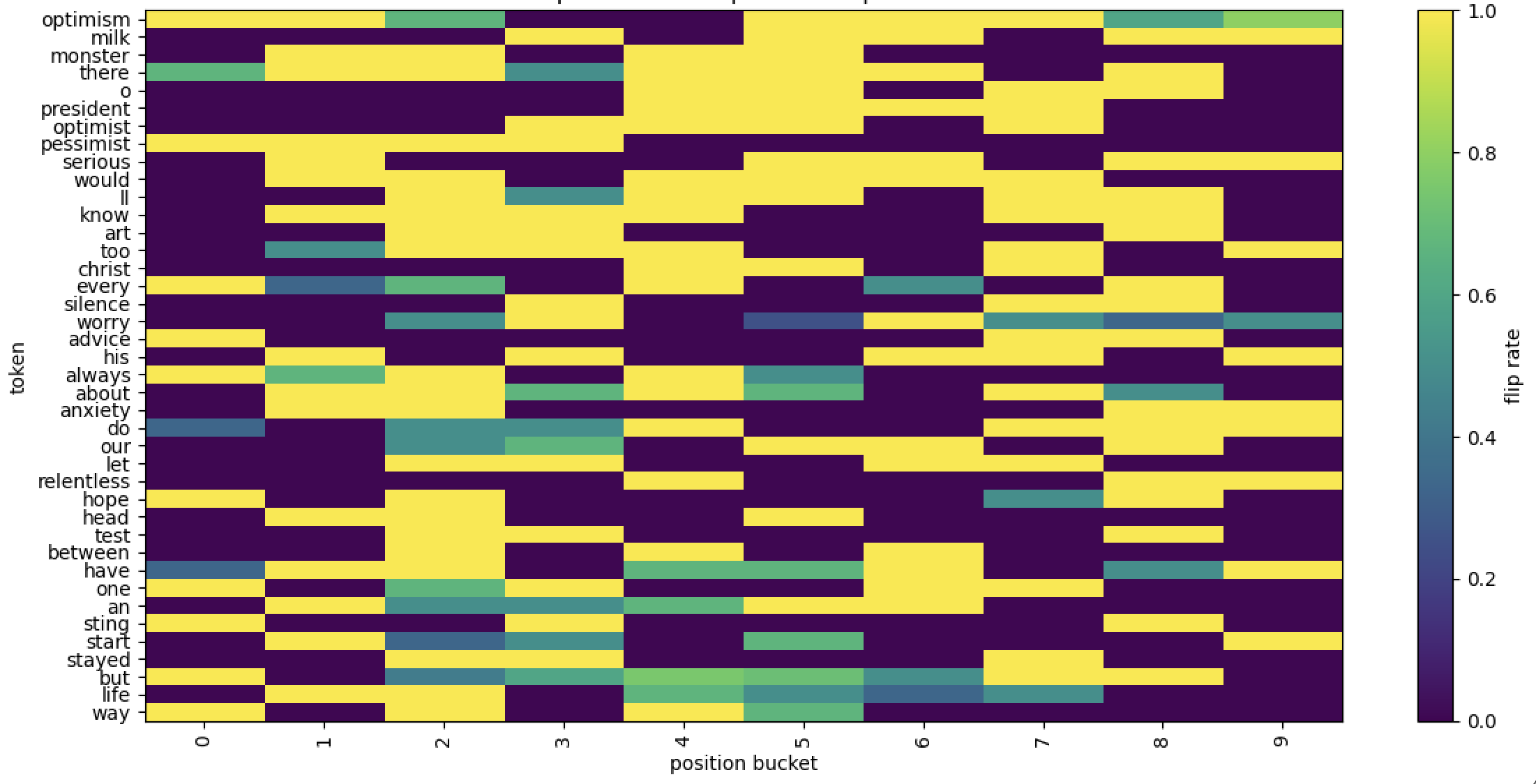}
\caption{Heat map of keywords by positional bucket for class \emph{Optimism}.}
\label{fig:optimism-heatmap}

\includegraphics[width=\textwidth]{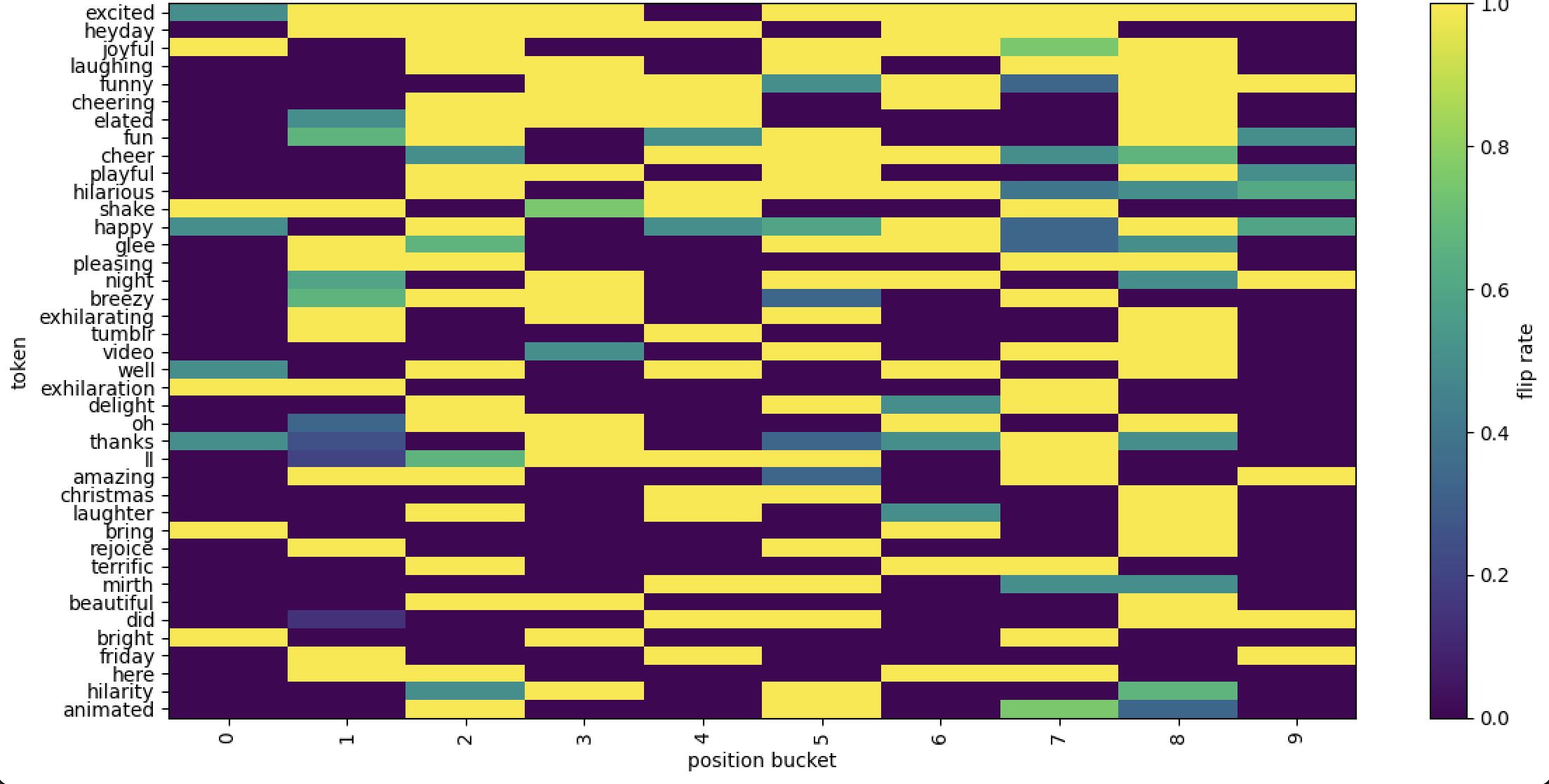}
\caption{Heat map of keywords by positional bucket for class \emph{Joy}.}
\label{fig:joy-heatmap}
\end{figure*}

Tables~\ref{fig:anger-heatmap}, \ref{fig:sadness-heatmap}, \ref{fig:optimism-heatmap}, and \ref{fig:joy-heatmap} present results for the classes \textit{Anger}, \textit{Sadness}, \textit{Optimism}, and \textit{Joy}, respectively. We identify \emph{causal tokens}—words whose masking flips the activation of at least one class-specific predicate neuron. These words are grouped into 10 buckets based on their relative position within the input.

\clearpage

\clearpage

\subsection{Top grounded rules}
Tables \ref{tab:layer1-rules}, \ref{tab:layer2-rules}, \ref{tab:layer3-rules}, \ref{tab:layer4-rules}, \ref{tab:layer5-rules}, \ref{tab:layer6-rules} present the top five grounded rules, i.e., (\textit{Keyword}, \textit{Template}) pairs, associated with the top five neurons in each corresponding layer, respectively. 

\begin{table}[H]
\centering
\small                  
\setlength{\tabcolsep}{4pt}   
\resizebox{\linewidth}{!}{%
\begin{tabular}{|r|l|r|l|l|r|r|}
\hline
\textbf{Neuron} & \textbf{Class} & \textbf{Purity} & \textbf{Keyword} & \textbf{Template} & \textbf{Flips} & \textbf{Rate} \\
\hline\hline
\multirow{5}{*}{438} & \multirow{5}{*}{optimism} & \multirow{5}{*}{1.2304} & is  & before\_verb   & 14 & 1.00 \\
                     &                            &                         & a   & exists         & 14 & 1.00 \\
                     &                            &                         & you & before\_verb   & 32 & 1.00 \\
                     &                            &                         & you & at\_start      & 25 & 1.00 \\
                     &                            &                         & you & after\_subject & 22 & 1.00 \\ \hline
\multirow{5}{*}{683} & \multirow{5}{*}{optimism} & \multirow{5}{*}{1.2237} & is  & before\_verb   & 17 & 1.00 \\
                     &                            &                         & a   & exists         & 13 & 1.00 \\
                     &                            &                         & you & before\_verb   & 21 & 1.00 \\
                     &                            &                         & you & at\_start      & 18 & 1.00 \\
                     &                            &                         & you & after\_subject & 16 & 1.00 \\ \hline
\multirow{5}{*}{568} & \multirow{5}{*}{optimism} & \multirow{5}{*}{1.2081} & is  & before\_verb   & 20 & 1.00 \\
                     &                            &                         & a   & exists         & 17 & 1.00 \\
                     &                            &                         & you & before\_verb   & 22 & 1.00 \\
                     &                            &                         & you & at\_start      & 22 & 1.00 \\
                     &                            &                         & you & after\_subject & 19 & 1.00 \\ \hline
\multirow{5}{*}{389} & \multirow{5}{*}{anger}    & \multirow{5}{*}{1.2037} & my  & at\_end        & 12 & 1.00 \\
                     &                            &                         & it  & at\_start      & 74 & 1.00 \\
                     &                            &                         & it  & exists         & 55 & 1.00 \\
                     &                            &                         & it  & before\_verb   & 52 & 1.00 \\
                     &                            &                         & it  & at\_end        & 43 & 1.00 \\ \hline
\multirow{5}{*}{757} & \multirow{5}{*}{optimism} & \multirow{5}{*}{1.2006} & is  & before\_verb   & 15 & 1.00 \\
                     &                            &                         & a   & exists         & 13 & 1.00 \\
                     &                            &                         & you & before\_verb   & 26 & 1.00 \\
                     &                            &                         & you & at\_start      & 23 & 1.00 \\
                     &                            &                         & you & after\_subject & 18 & 1.00 \\ \hline
\end{tabular}}
\caption{Top-5 grounded rules in layer~1.}
\label{tab:layer1-rules}
\end{table}

\begin{table}[H]
\centering
\small                   
\setlength{\tabcolsep}{4pt}  
\resizebox{\linewidth}{!}{%
\begin{tabular}{|r|l|r|l|l|r|r|}
\hline
\textbf{Neuron} & \textbf{Class} & \textbf{Purity} & \textbf{Keyword} & \textbf{Template} & \textbf{Flips} & \textbf{Rate} \\
\hline\hline
\multirow{5}{*}{734} & \multirow{5}{*}{anger} & \multirow{5}{*}{1.1921} & it   & at\_start      & 61 & 1.00 \\
                     &                        &                         & it   & after\_verb    & 73 & 1.00 \\
                     &                        &                         & s    & after\_subject & 62 & 1.00 \\
                     &                        &                         & that & after\_subject & 84 & 1.00 \\
                     &                        &                         & that & after\_verb    & 87 & 1.00 \\ \hline
\multirow{5}{*}{110} & \multirow{5}{*}{anger} & \multirow{5}{*}{1.1875} & it   & at\_start      &  6 & 1.00 \\
                     &                        &                         & it   & after\_verb    &  6 & 1.00 \\
                     &                        &                         & s    & after\_subject &  5 & 1.00 \\
                     &                        &                         & that & after\_subject &  3 & 1.00 \\
                     &                        &                         & that & after\_verb    &  1 & 1.00 \\ \hline
\multirow{5}{*}{756} & \multirow{5}{*}{anger} & \multirow{5}{*}{1.1739} & it   & at\_start      & 52 & 1.00 \\
                     &                        &                         & it   & after\_verb    & 63 & 1.00 \\
                     &                        &                         & s    & after\_subject & 61 & 1.00 \\
                     &                        &                         & that & after\_subject & 64 & 1.00 \\
                     &                        &                         & that & after\_verb    & 70 & 1.00 \\ \hline
\multirow{5}{*}{635} & \multirow{5}{*}{anger} & \multirow{5}{*}{1.1722} & it   & at\_start      & 53 & 1.00 \\
                     &                        &                         & it   & after\_verb    & 60 & 1.00 \\
                     &                        &                         & s    & after\_subject & 65 & 1.00 \\
                     &                        &                         & that & after\_subject & 69 & 1.00 \\
                     &                        &                         & that & after\_verb    & 78 & 1.00 \\ \hline
\multirow{5}{*}{453} & \multirow{5}{*}{anger} & \multirow{5}{*}{1.1628} & it   & at\_start      & 52 & 1.00 \\
                     &                        &                         & it   & after\_verb    & 62 & 1.00 \\
                     &                        &                         & s    & after\_subject & 64 & 1.00 \\
                     &                        &                         & that & after\_subject & 69 & 1.00 \\
                     &                        &                         & that & after\_verb    & 74 & 1.00 \\ \hline
\end{tabular}}
\caption{Top-5 grounded rules in layer~2.}
\label{tab:layer2-rules}
\end{table}

\begin{table}[H]
\centering
\small         
\setlength{\tabcolsep}{4pt} 
\resizebox{\linewidth}{!}{%
\begin{tabular}{|r|l|r|l|l|r|r|}
\hline
\textbf{Neuron} & \textbf{Class} & \textbf{Purity} & \textbf{Keyword} & \textbf{Template} & \textbf{Flips} & \textbf{Rate} \\
\hline\hline
\multirow{1}{*}{597} & \multirow{1}{*}{joy}   & \multirow{1}{*}{1.4149} & amazing   & after\_verb     &  1 & 1.00 \\ \hline
\multirow{5}{*}{509} & \multirow{5}{*}{anger} & \multirow{5}{*}{1.3988} & it        & at\_end         & 16 & 1.00 \\
                     &                        &                         & that      & before\_subject &  8 & 1.00 \\
                     &                        &                         & he        & before\_verb    & 23 & 1.00 \\
                     &                        &                         & he        & at\_start       & 19 & 1.00 \\
                     &                        &                         & user      & after\_subject  & 16 & 1.00 \\ \hline
\multirow{5}{*}{495} & \multirow{5}{*}{anger} & \multirow{5}{*}{1.3886} & it        & at\_end         & 10 & 1.00 \\
                     &                        &                         & that      & before\_subject & 14 & 1.00 \\
                     &                        &                         & he        & before\_verb    & 10 & 1.00 \\
                     &                        &                         & he        & at\_start       & 12 & 1.00 \\
                     &                        &                         & user      & after\_subject  & 13 & 1.00 \\ \hline
\multirow{5}{*}{652} & \multirow{5}{*}{joy}   & \multirow{5}{*}{1.3743} & amazing   & after\_verb     & 28 & 1.00 \\
                     &                        &                         & live      & after\_verb     & 27 & 1.00 \\
                     &                        &                         & ly        & before\_verb    & 27 & 1.00 \\
                     &                        &                         & ly        & at\_start       & 27 & 1.00 \\
                     &                        &                         & musically & at\_end         & 27 & 1.00 \\ \hline
\multirow{5}{*}{734} & \multirow{5}{*}{anger} & \multirow{5}{*}{1.3660} & it        & at\_end         &  6 & 1.00 \\
                     &                        &                         & that      & before\_subject &  2 & 1.00 \\
                     &                        &                         & he        & before\_verb    &  1 & 1.00 \\
                     &                        &                         & he        & at\_start       &  1 & 1.00 \\
                     &                        &                         & user      & after\_subject  &  3 & 1.00 \\ \hline
\end{tabular}}
\caption{Top-5 grounded rules in layer~3.}
\label{tab:layer3-rules}
\end{table}

\begin{table}[h]
\centering
\small
\setlength{\tabcolsep}{4pt}          
\resizebox{\linewidth}{!}{
\begin{tabular}{|r|l|r|l|l|r|r|}
\hline
\textbf{Neuron} & \textbf{Class} & \textbf{Purity} & \textbf{Keyword} & \textbf{Template} & \textbf{Flips} & \textbf{Rate} \\
\hline\hline
\multirow{1}{*}{597} & \multirow{1}{*}{joy}   & \multirow{1}{*}{1.5896} & is        & before\_verb    &  1 & 1.00 \\ \hline
\multirow{5}{*}{232} & \multirow{5}{*}{joy}   & \multirow{5}{*}{1.5464} & is        & before\_verb    &  4 & 1.00 \\
                     &                        &                          & amazing   & after\_verb     &  1 & 1.00 \\
                     &                        &                          & \textit{i}& after\_subject  &  6 & 0.96 \\
                     &                        &                          & \textit{i}& after\_verb     &  8 & 0.96 \\
                     &                        &                          & this      & after\_verb     &  5 & 0.96 \\ \hline
\multirow{2}{*}{66}  & \multirow{2}{*}{joy}   & \multirow{2}{*}{1.5405} & is        & before\_verb    &  2 & 1.00 \\
                     &                        &                          & this      & after\_verb     &  1 & 0.96 \\ \hline
\multirow{5}{*}{399} & \multirow{5}{*}{anger} & \multirow{5}{*}{1.5323} & s         & exists          &  2 & 1.00 \\
                     &                        &                          & fucking   & after\_subject  &  4 & 1.00 \\
                     &                        &                          & but       & before\_subject &  1 & 1.00 \\
                     &                        &                          & but       & at\_start       &  1 & 1.00 \\
                     &                        &                          & but       & before\_verb    &  1 & 1.00 \\ \hline
\multirow{5}{*}{71}  & \multirow{5}{*}{joy}   & \multirow{5}{*}{1.5262} & is        & before\_verb    &  8 & 1.00 \\
                     &                        &                          & amazing   & after\_verb     &  1 & 1.00 \\
                     &                        &                          & \textit{i}& after\_subject  & 20 & 0.96 \\
                     &                        &                          & \textit{i}& after\_verb     & 17 & 0.96 \\
                     &                        &                          & this      & after\_verb     &  9 & 0.96 \\ \hline
\end{tabular}}
\caption{Top-5 grounded rules in layer~4.}
\label{tab:layer4-rules}
\end{table}
\begin{table}[H]
\centering
\small
\setlength{\tabcolsep}{4pt}
\resizebox{\linewidth}{!}{%
\begin{tabular}{|r|l|r|l|l|r|r|}
\hline
\textbf{Neuron} & \textbf{Class} & \textbf{Purity} & \textbf{Keyword} & \textbf{Template} & \textbf{Flips} & \textbf{Rate} \\
\hline\hline
\multirow{1}{*}{499} & \multirow{1}{*}{joy} & \multirow{1}{*}{1.8198} & today   & at\_start      & 1 & 1.00 \\ \hline
\multirow{1}{*}{258} & \multirow{1}{*}{joy} & \multirow{1}{*}{1.7694} & heyday  & after\_verb    & 1 & 1.00 \\ \hline
\multirow{2}{*}{698} & \multirow{2}{*}{joy} & \multirow{2}{*}{1.7653} & heyday  & after\_verb    & 1 & 1.00 \\
                     &                      &                          & today   & at\_start      & 3 & 1.00 \\ \hline
\multirow{1}{*}{221} & \multirow{1}{*}{joy} & \multirow{1}{*}{1.7426} & today   & at\_start      & 1 & 1.00 \\ \hline
\multirow{3}{*}{535} & \multirow{3}{*}{joy} & \multirow{3}{*}{1.7384} & heyday  & after\_verb    & 2 & 1.00 \\
                     &                      &                          & glee    & before\_subject& 6 & 1.00 \\
                     &                      &                          & today   & at\_start      & 2 & 1.00 \\ \hline
\end{tabular}}
\caption{Top-5 grounded rules in layer~5.}
\label{tab:layer5-rules}
\end{table}

\begin{table}[H]
\centering
\small
\setlength{\tabcolsep}{4pt}          
\resizebox{\linewidth}{!}{%
\begin{tabular}{|r|l|r|l|l|r|r|}
\hline
\textbf{Neuron} & \textbf{Class} & \textbf{Purity} & \textbf{Keyword} & \textbf{Template} & \textbf{Flips} & \textbf{Rate} \\
\hline\hline
\multirow{2}{*}{122} & \multirow{2}{*}{joy} & \multirow{2}{*}{1.8478} & laughter  & after\_subject & 2 & 1.00 \\
                     &                      &                         & hilarious & after\_subject & 1 & 0.88 \\ \hline
\multirow{5}{*}{344} & \multirow{5}{*}{joy} & \multirow{5}{*}{1.8359} & playful   & exists         & 1 & 1.00 \\
                     &                      &                         & smiling   & after\_subject & 2 & 1.00 \\
                     &                      &                         & laughter  & at\_end        & 1 & 1.00 \\
                     &                      &                         & laughter  & after\_subject & 1 & 1.00 \\
                     &                      &                         & hilarious & after\_subject & 2 & 0.88 \\ \hline
\multirow{2}{*}{497} & \multirow{2}{*}{joy} & \multirow{2}{*}{1.8342} & laughter  & after\_subject & 1 & 1.00 \\
                     &                      &                         & hilarious & after\_subject & 1 & 0.88 \\ \hline
\multirow{5}{*}{212} & \multirow{5}{*}{joy} & \multirow{5}{*}{1.8330} & playful   & exists         & 1 & 1.00 \\
                     &                      &                         & omg       & before\_subject& 1 & 1.00 \\
                     &                      &                         & smiling   & after\_subject & 2 & 1.00 \\
                     &                      &                         & laughter  & at\_end        & 1 & 1.00 \\
                     &                      &                         & laughter  & after\_subject & 3 & 1.00 \\ \hline
\multirow{2}{*}{452} & \multirow{2}{*}{joy} & \multirow{2}{*}{1.8261} & laughter  & after\_subject & 1 & 1.00 \\
                     &                      &                         & hilarious & after\_subject & 1 & 0.88 \\ \hline
\end{tabular}}
\caption{Top-5 grounded rules in layer~6.}
\label{tab:layer6-rules}
\end{table}

\clearpage
\subsection{Top three rules per class for every layer}
Tables~\ref{tab:top3-l123} and~\ref{tab:top3-l456} present the top three grounded rules for each class across all layers.

\begin{table}[h]
\centering
\small
\setlength{\tabcolsep}{4pt}\renewcommand{\arraystretch}{1.1}
\resizebox{\columnwidth}{!}{%
\begin{tabular}{|l|l|l|r|r|r|r|}
\hline
\textbf{Class}&\textbf{Keyword} & \textbf{Template}&\textbf{Flips}&\textbf{Total}&\textbf{Rate}&\textbf{Neuron}\\\hline\hline
\multicolumn{7}{|c|}{\textbf{Layer 1}}\\\hline
anger&terrorism&after\_subject&19&19&1.00&125\\
anger&am&at\_start&19&19&1.00&125\\
anger&terrorism&after\_subject&17&19&1.00&695\\
joy&ly&before\_verb&27&27&1.00&505\\
joy&ly&at\_start&27&27&1.00&505\\
joy&musically&at\_end&27&27&1.00&505\\
optimism&can&after\_subject&17&19&1.00&563\\
optimism&your&after\_verb&18&21&1.00&438\\
optimism&your&after\_verb&18&21&1.00&563\\
sadness&want&after\_subject&17&19&1.00&52\\
sadness&lost&after\_subject&24&30&1.00&52\\
sadness&want&after\_subject&16&19&1.00&679\\\hline
\multicolumn{7}{|c|}{\textbf{Layer 2}}\\\hline
anger&have&after\_subject&67&86&1.00&298\\
anger&with&after\_verb&67&84&1.00&298\\
anger&have&after\_subject&65&86&1.00&136\\
sadness&my&after\_verb&83&94&1.00&698\\
sadness&is&after\_subject&96&110&1.00&712\\
sadness&my&after\_verb&72&94&1.00&712\\\hline
\multicolumn{7}{|c|}{\textbf{Layer 3}}\\\hline
anger&people&before\_verb&27&32&1.00&189\\
anger&why&before\_verb&23&30&1.00&189\\
anger&why&before\_subject&23&31&1.00&189\\
joy&ly&before\_verb&27&27&1.00&652\\
joy&ly&before\_verb&27&27&1.00&28\\
joy&ly&at\_start&27&27&1.00&652\\
optimism&be&after\_subject&22&26&1.00&459\\
optimism&be&after\_subject&22&26&1.00&416\\
optimism&be&after\_subject&19&26&1.00&157\\
sadness&sad&at\_end&25&27&1.00&316\\
sadness&sad&at\_end&24&27&1.00&498\\
sadness&sad&after\_verb&23&26&1.00&305\\\hline
\end{tabular}}
\caption{Top-3 grounded rules per class (Layers 1 to 3).}
\label{tab:top3-l123}
\end{table}

\newpage
\begin{table}[h]
\centering
\small
\setlength{\tabcolsep}{4pt}\renewcommand{\arraystretch}{1.1}
\resizebox{\columnwidth}{!}{%
\begin{tabular}{|l|l|l|r|r|r|r|}
\hline
\textbf{Class}&\textbf{Token}&\textbf{Keyword} & \textbf{Template}&\textbf{Total}&\textbf{Rate}&\textbf{Neuron}\\\hline\hline
\multicolumn{7}{|c|}{\textbf{Layer 4}}\\\hline
anger&fucking&after\_subject&20&20&1.00&434\\
anger&anger&after\_verb&18&19&1.00&92\\
anger&terrorism&after\_subject&17&19&1.00&92\\
joy&amazing&after\_verb&29&31&1.00&95\\
joy&this&after\_verb&41&48&0.96&95\\
joy&is&before\_verb&19&25&1.00&95\\
optimism&be&after\_subject&22&26&1.00&416\\
optimism&you&at\_start&26&34&0.97&416\\
optimism&you&before\_verb&29&38&1.00&416\\
sadness&sad&at\_end&19&27&1.00&305\\
sadness&sad&at\_end&19&27&1.00&23\\
sadness&sad&at\_end&17&27&1.00&296\\\hline
\multicolumn{7}{|c|}{\textbf{Layer 5}}\\\hline
anger&awful&at\_end&15&19&0.84&92\\
anger&angry&after\_subject&22&27&0.93&92\\
anger&angry&after\_verb&21&27&0.89&92\\
optimism&it&at\_start&9&24&0.83&430\\
optimism&it&before\_verb&6&21&0.81&430\\
optimism&is&at\_start&8&29&0.76&459\\
sadness&sad&at\_end&23&27&0.93&246\\
sadness&sadness&exists&17&23&0.83&433\\
sadness&sad&at\_end&22&27&0.93&433\\\hline
\multicolumn{7}{|c|}{\textbf{Layer 6}}\\\hline
anger&anger&after\_verb&14&19&0.89&531\\
anger&awful&at\_end&13&19&0.74&15\\
anger&anger&after\_verb&13&19&0.89&603\\
optimism&not&after\_subject&14&19&0.74&142\\
optimism&s&after\_subject&13&20&0.75&142\\
optimism&user&exists&15&20&0.80&142\\
sadness&sadness&exists&17&23&0.78&298\\
sadness&sad&exists&25&31&0.81&298\\
sadness&sad&at\_end&21&27&0.89&242\\\hline
\end{tabular}}
\caption{Top-3 grounded rules per class (Layers 4 to 6).}
\label{tab:top3-l456}
\end{table}

\end{document}